\documentclass[sigconf]{acmart}
\usepackage{graphicx}
\usepackage{balance}  
\usepackage{graphics}
\usepackage{mdwlist}
\usepackage{float}
\usepackage{algorithm}
\usepackage{hyperref}
\usepackage{subcaption}
\usepackage{rotating}
\usepackage{xcolor}
\usepackage{tabularx}
\hypersetup{
    colorlinks=true,
    linkcolor=blue,
    filecolor=magenta,      
    urlcolor=cyan,
}

\copyrightyear{2025}
\acmYear{2025}
\setcopyright{cc}
\setcctype{by}
\acmConference[CHI '25]{CHI Conference on Human Factors in Computing Systems}{April 26-May 1, 2025}{Yokohama, Japan}
\acmBooktitle{CHI Conference on Human Factors in Computing Systems (CHI '25), April 26-May 1, 2025, Yokohama, Japan}\acmDOI{10.1145/3706598.3713181}
\acmISBN{979-8-4007-1394-1/25/04}

\begin{document}

\title{Think Together and Work Better: Combining Humans' and LLMs' Think-Aloud Outcomes for Effective Text Evaluation}

\author{Seong Yeub Chu}
\authornote{Both authors contributed equally to this research.}
\affiliation{%
  \institution{Graduate School of Data Science, KAIST}
  \city{Daejeon}
  \country{Republic of Korea}
}
\email{chseye7@kaist.ac.kr}

\author{Jong Woo Kim}
\authornotemark[1]
\affiliation{%
  \institution{Department of Industrial \& Systems Engineering, KAIST}
  \city{Daejeon}
  \country{Republic of Korea}
}
\email{gsds4885@kaist.ac.kr}

\author{Mun Yong Yi}
\authornote{Corresponding author}
\affiliation{%
  \institution{Graduate School of Data Science, KAIST}
  \city{Daejeon}
  \country{Republic of Korea}
}
\email{munyi@kaist.ac.kr}

\begin{abstract}
    This study introduces \textbf{InteractEval}, a framework that integrates the outcomes of Think-Aloud (TA) conducted by humans and LLMs to generate attributes for checklist-based text evaluation. By combining humans' flexibility and high-level reasoning with LLMs' consistency and extensive knowledge, InteractEval outperforms text evaluation baselines on a text summarization benchmark (SummEval) and an essay scoring benchmark (ELLIPSE). Furthermore, an in-depth analysis shows that it promotes divergent thinking in both humans and LLMs, leading to the generation of a wider range of relevant attributes and enhancement of text evaluation performance. A subsequent comparative analysis reveals that humans excel at identifying attributes related to internal quality (Coherence and Fluency), but LLMs perform better at those attributes related to external alignment (Consistency and Relevance). Consequently, leveraging both humans and LLMs together produces the best evaluation outcomes, highlighting the necessity of effectively combining humans and LLMs in an automated checklist-based text evaluation.
\end{abstract}

\renewcommand{\shortauthors}{SY. Chu, JW. Kim, \& MY. Yi}

\ccsdesc[500]{Human-centered computing~Human computer interaction (HCI)}

\keywords{Large Language Model, Think-Aloud, Human-LLM Combination, Text Evaluation, Checklist}

\begin{teaserfigure}
  \centering
  \includegraphics[height=7.7cm, width=14.7cm]{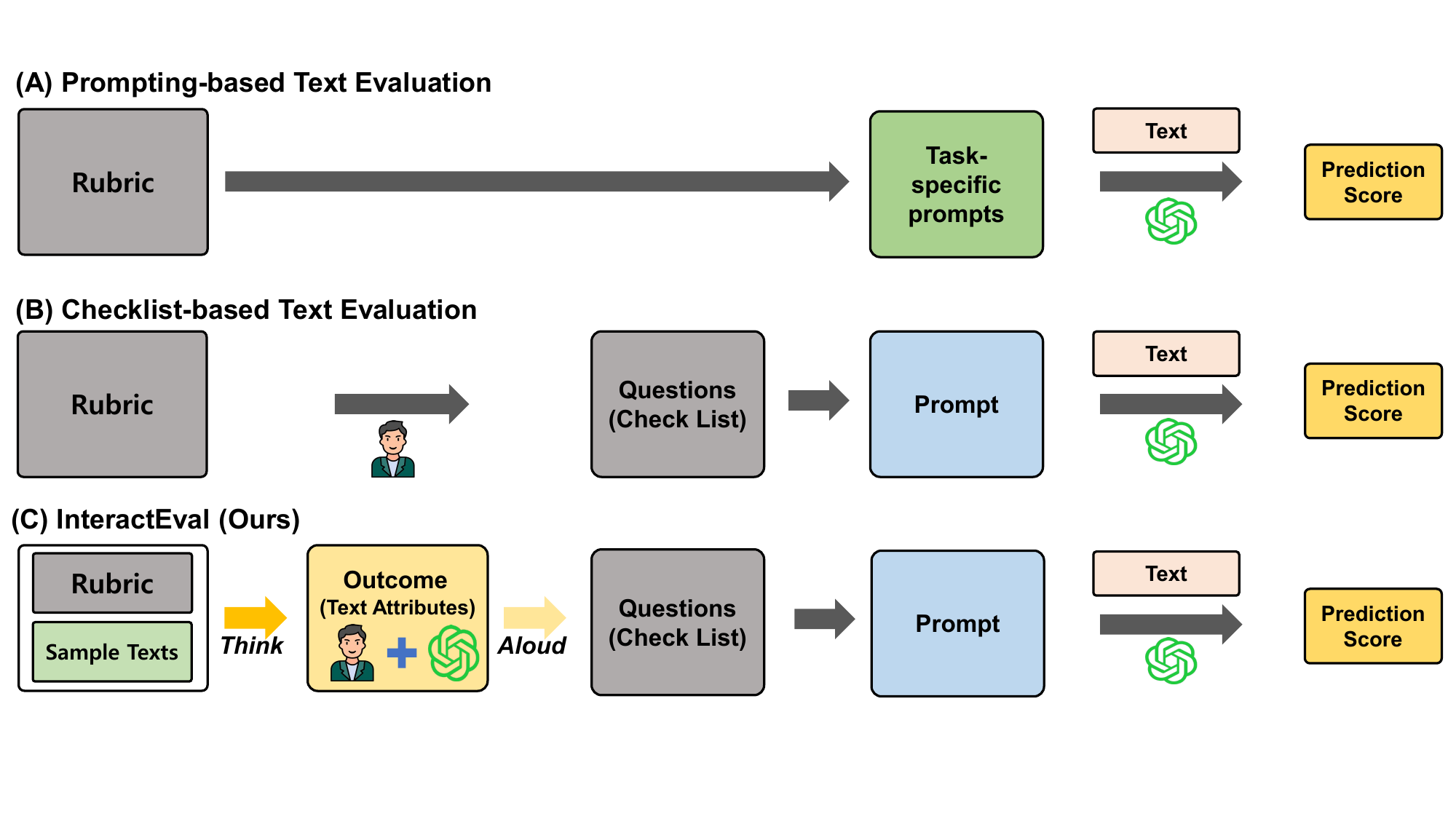}
  \caption{Comparative Overview of Prompt-based, Checklist-based, and the Proposed InteractEval Text Evaluation Methods: (A) Prompt-based evaluation generates task-specific prompts from a rubric for prediction scores. (B) Checklist-based evaluation uses human-created questions from the rubric to form prompts. (C) InteractEval combines human and LLM-generated attributes using Think Aloud methods to create questions and produce final prediction scores.}
  \label{fig1}
\end{teaserfigure}

\maketitle

\section{Introduction}
\label{intro}
In recent years, the rapid advancement of large language models (LLMs) has led to a surge in research focused on text generation \cite{lee, pearl, peer, yang, lee2024}, not to mention the industries: Notion AI \footnote{https://www.notion.so/product}, Jasper \footnote{https://www.jasper.ai/}, Writesonic \footnote{https://writesonic.com/}, sudoWrite \footnote{https://www.sudowrite.com/}, and Cohesive \footnote{https://cohesive.so/}, to name a few textual content generation solutions. Accordingly, the need for accurate mechanisms to evaluate the quality of automatically generated text has grown increasingly critical \cite{gptscore, geval, checkeval, wang}, and led to various LLM-based evaluation approaches. LLM-based evaluation methods have introduced significant improvements by enabling reference-free evaluations using the chain-of-thought (CoT) prompting strategy \cite{geval} and adjusting scores based on output token probabilities \cite{gptscore}. However, as shown in Figure \ref{fig1} (A), these methods depend heavily on task-specific prompts and tend to only evaluate the general aspects of the summary, leading to reduced reliability. To address these challenges, checklist-based approaches \cite{checkeval, lee2024} introduce the use of checklists to evaluate text by breaking down rating dimensions into several key components. These methods attempt to enhance the accuracy of the evaluation by having the LLM focus on specific components when scoring and analyzing the summary, thereby assessing the summary based on detailed elements of the dimension. 

Nevertheless, concerns persist regarding the manually created key components in these checklists. As shown in Figure \ref{fig1} (B), the checklists are generated by LLMs through sub-questions derived from key components manually defined by the authors based on a rubric. This process raises two main concerns. First, the reliance on human-defined key components introduces a strong dependency on human judgment. According to prior studies, humans and models each possess distinct strengths and weaknesses \cite{pantcheva, blattberg, wang2024, das}. Thus, depending exclusively on human-defined key components may inherit the limitations of human abilities in the evaluation process. Secondly, it raises the question of whether these components fully capture how the aspects intended to be evaluated in the rubrics' rating dimensions manifest in the actual text. This approach follows a theory-based method \cite{plakans, rodriguez}, where a rating scale is designed based on established theories or pre-defined guidelines, such as rubrics. This method potentially overlooks the detailed features represented in example texts, focusing only on the rubric, and bring about a coarse-grained checklists \cite{brindley, turner}.

Our study introduces \textbf{InteractEval}, addressing the limitations of relying solely on either humans or machines by combining their ideas to mitigate shortcomings that arise from over-reliance on one source. We integrate the ideas of LLMs and humans derived from think-aloud (TA) processes, which have been recognized for fostering divergent thinking \cite{ericsson1, ericsson2, nielsen, lewis}. During the TA procedure, we merge performance-driven approaches, which review individual samples, with the theory-based methods, inducing humans and LLMs to specifically consider how the aspects intended to be rated in a dimension are reflected in sample texts \cite{fulcher, knoch, jeffrey}. These combined TA results are then used to create unbiased and fine-grained checklists for text evaluation as illustrated in Figure \ref{fig1} (C). 

We expect the human-LLM collaboration to complement their distinct advantages via the combination \cite{geval, wang2024}. Specifically, humans often struggle with inconsistent performance when handling large tasks, which can significantly hinder their performance \cite{li2020impact}. In text evaluation, humans often find it challenging to identify relevance between two documents due to the need for consistent reading of large volumes of text. On the other hand, automated models heavily rely on extensive datasets \cite{wang2024, das}, which restrict them to the knowledge and patterns explicitly present in the training data, often making it difficult for them to fully capture the complex semantic details of a given text. For example, models may encounter difficulties to identify coherent and hierarchical structure in a given text \cite{pantcheva}, highlighting the models' vulnerability to the quality and diversity of the data they are trained on. Therefore, integrating the consistent performance of models with humans' flexibility, prior knowledge, and high-level reasoning capability (e.g., critical thinking, creativity, imagination, diverse perspective-taking, situational awareness, and intuitive thinking) can enhance the reliability of text evaluation by producing results more closely aligned with human judgement \cite{blattberg, das, scholastic, wu2023reasoning, pantcheva}.

Our framework contains three main steps to combine the outcomes of TA and create checklists based on them. First, after the TA produres conducted by humans and LLMs, we collect the results from both of them, referred to as text attributes, which are needed for rating specific dimensions of texts. Next, we prompt an LLM to extract key components from the merged ideas, to cluster the merged attributes based on these components, and to generate questions related to each component. Lastly, we employ the LLM to create a checklist for rating a specific dimension of the given texts. 

This study examines whether combining TA results from both humans and LLMs improves text evaluation performance. We conduct extensive experiments to test whether the TA method we use enables humans and LLMs to generate diverse attributes that contribute to building a refined checklist and enhancing evaluation quality. Additionally, we explore how the checklist, based on these combined attributes, further improves text evaluation. In sum, our study examines three key research questions, as outlined below:

\renewcommand{\labelenumi}{\textbf{RQ} 
\textbf{\arabic{enumi}}.}
\begin{enumerate}
    \item\label{rq1} Does our text evaluation framework, which uses checklists designed with integrated attributes from human experts and LLMs through TA, outperform baseline text evaluation methods in terms of correlation with ground-truth scores?
    \item\label{rq2} Do the attributes generated through TA improve the evaluation performance of generated texts, based on the checklists constructed from those attributes?
    \item\label{rq3}How effective is the combination of attributes from humans' TA and LLMs' TA for text evaluation using the checklists constructed from the TA-generated attributes?
\end{enumerate}

\section{Related Work}
\label{related_work}
\subsection{Text Evaluation with Large Language Models}

LLM-based text evaluation has emerged to address the shortcomings of traditional text evaluation methods, such as BLUE \cite{bleu} or ROUGE-L \cite{rouge}, which highly depend on reference text \cite{summeval, gu, guan}. Research on text evaluation using LLMs \cite{geval, gptscore, kocmi, chateval, debate} has been actively progressing, starting with GPTScore \cite{gptscore}, which scores texts based on its token generation probability, and G-Eval \cite{geval}, which uses the Chain-of-Thought (CoT) \cite{cot} technique to evaluate texts. Studies utilizing multi-agent collaboration \cite{chateval, debate} are particularly active, with the aim of overcoming the single-thread limitations \cite{moa} of single-LLM approaches. The single-LLM methods hinder scalability and speed when dealing with complex tasks like text evaluation, which requires taking multiple aspects of text quality into consideration. Despite these advantages, those multi-agent based methods are resource-intensive, incurring high costs in time and computation. They experience increased time to first token (TTFT) \cite{moa}, and still fall short in their robustness on text evaluation. CheckEval \cite{checkeval}, which are particularly effective for yes/no questions, introduced a binary evaluation method using checklists to assess texts, based on the observation that LLMs excel at fine-grained tasks \cite{liu, min}. However, its heavy reliance on humans (specifically the authors) in designing key components for checklist creation raises concerns about the reliability of the evaluation criteria \cite{checkeval}. Moreover, it still suffers from being coarse-grained by designing the key components solely based on the broad rubric definition of each dimension.

\subsection{Combination of Human Intelligence and Artificial Intelligence}
Human experts and models each have unique strengths and weaknesses, making them complementary when used together. Humans are more flexible but can become fatigued or bored with large tasks, while models are more consistent and rigid \cite{blattberg}. Given the substantial prior knowledge (e.g. universal grammar \cite{universal, chomsky}) humans possess, many studies in the human-in-the-loop (HITL) field aim to enhance the consistent performance of neural models by incorporating human expertise into data processing, model training, inference, and system development \cite{wu2022, xin}. Human elements, such as empirical knowledge and practical skills, help models learn and perform specific tasks, especially when dealing with sparse data \cite{wu2022, kumar}. On the other hand, models are capable of managing demanding workloads efficiently, although they heavily rely on datasets. Specifically, LLMs can analyze vast text corpora of academic papers to find a new research insight \cite{oppenlaender}. Additionally, incorporating LLMs into the research cycle accelerates the overall process and enhances efficiency. However, this approach may face challenges such as hallucinated outputs, misalignment with established referent knowledge, and self-contradictions \cite{elagroudy}. LLM performance, notably, is highly dependent on specific tasks, datasets, and labels \cite{zhu, ziems}, and they often retain biases and errors in their outputs \cite{geval, wang2024}, making human oversight crucial. 

Numerous studies have demonstrated that combining human and AI for decision-making outperforms decisions in isolation \cite{blattberg, wang2024, das, revilla, cohen}. Various studies have explored the effectiveness of collaboration between humans and AI models across various tasks, including text generation and non-text object generation, but its potential in text evaluation remains underexamined. For example, synergistic collaboration between humans and LLMs, leveraging human cognitive strengths, produces higher-quality outputs in co-writing \cite{dhillon}. Additionally, a machine-in-the-loop approach can enhance the workflow of human analyst's text analysis by utilizing human knowledge as a foundation \cite{scholastic}. Highly refined prompt templates, developed through the aggregation of diverse human expertise, can be employed to make LLMs precisely audit their own outputs \cite{rastogi}, and collaboration between humans and LLMs supports the creative processes of individual designers or design teams for generating products \cite{zhou2024, han2024teams}. Furthermore, humans and LLMs can cooperate to improve the accuracy of data annotation \cite{wang2024}. Because most LLMs are trained with human feedback \cite{gpt4, llama, claude}, and tasks like text generation and summarization can be improved with HITL \cite{ziegler, stiennon}, involving humans in checklist construction should enhance the quality of LLMs' text evaluation.

\subsection{Think Aloud}
The think aloud (TA) method, where participants verbalize their thoughts while performing experimental tasks, is a well-established technique in cognitive psychology and usability studies. Introduced by Ericsson and Simon \cite{ericsson1}, this method has been pivotal in understanding cognitive processes and identifying usability issues by capturing real-time user feedback \cite{nielsen, lewis, concurrent}. There are two main types in the TA method: concurrent TA (CTA) and retrospective TA (RTA) \cite{ericsson2}. In CTA, users verbalize their thoughts while performing a task, whereas in RTA, they recall and describe their thoughts and feelings after completing the task. Although both methods have their own strengths, they also have limitations \cite{alhadreti}. 

The TA method has been widely applied beyond usability studies, serving as a tool across various fields to develop new methodologies by capturing diverse ideas \cite{ericsson1, nielsen}. For example, it was used to construct a rating scale for English writing assessment by having in-service teachers and experts think aloud their thoughts on relevant text attributes while reviewing sample essays and scoring dimensions \cite{wu}. These insights were then consolidated into a detailed rubric, providing a structured approach to rubric creation. Building upon the prior work, our study seeks to create a fine-grained checklist for evaluating texts. Using the TA method, we involve both humans and LLMs to identify specific text attributes for the rating dimensions - Coherence, Fluency, Consistency, and Relevance - ultimately refining and improving LLM-based evaluations.

\section{Method}
\label{method}
\begin{figure*}[t]
    \centering
    \includegraphics[width=\linewidth]{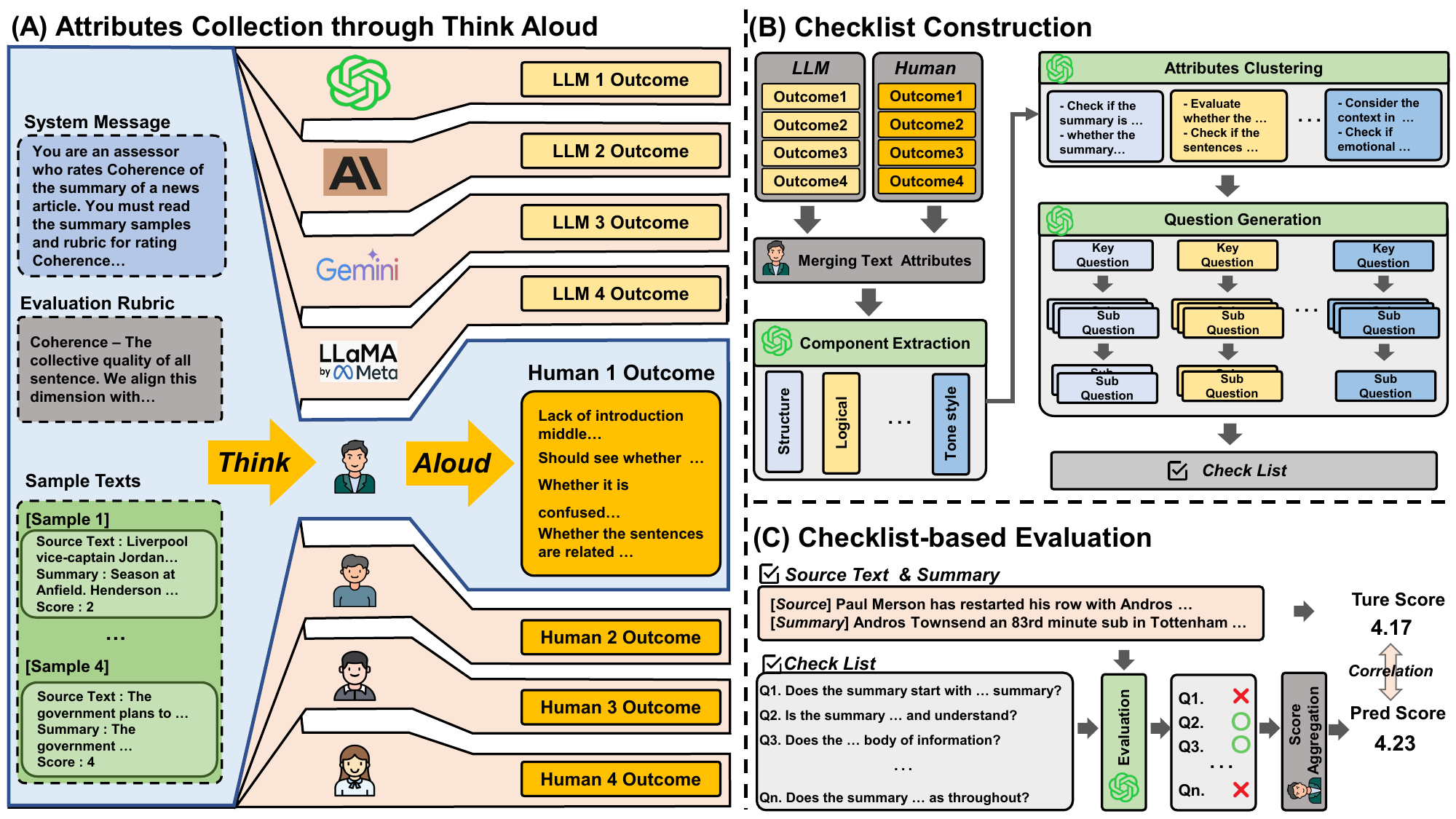}
    \caption{InteractEval Framework for Text Evaluation: \textbf{(A) Think Aloud}: Human experts verbalize their thoughts and LLMs articulate their knowledge to generate text attributes insights using sample texts and evaluation rubrics. \textbf{(B) Checklist Construction}: Insights are combined and categorized into key components, leading to the generation and validation of checklist questions. \textbf{(C) Checklist-based Evaluation}: The checklists are answered by an evaluator LLM to evaluate the summaries, with results aggregated into a final score, which is then checked against a ground-truth score.}
    \label{fig2}
\end{figure*}

\subsection{Overview}
InteractEval combines the TA outcomes from humans and LLMs to build checklists for reliable text evaluation. Inspired by the competitive and robust performance of CheckEval \cite{checkeval} in text evaluation, this study adopts a checklist-based evaluation approach. InteractEval consists of three distinct stages, as illustrated in Figure \ref{fig2}. Initially, humans participate in the TA process, while LLMs are prompted in a manner identical to TA, as shown in Figure \ref{fig3}, independently producing the text attributes to be considered when rating texts. These attributes are then merged and used to create checklists consisting of multiple boolean-style questions, requiring binary responses (Yes/No). To rigorously and efficiently assess the impact of human-AI collaboration, we limit its scope to the TA stage, specifically evaluating how collaborative attributes generated during TA contribute to the checklist-based evaluation framework. The evaluator LLM reviews the text summaries based on the checklist and provides a response to each question. Finally, the "Yes" responses are tallied to generate a final score, which is then compared with the human-assigned ground truth scores. A key distinction from CheckEval is that InteractEval generates the components using the attributes generated by both human and LLM, reducing shortcomings associated with each source and producing more fine-grained criteria \cite{pantcheva, blattberg, brindley, turner}.

\subsection{Dimensions for Evaluation}
Our study uses SummEval \cite{summeval} dataset, which includes a collection of news article summaries, for the main case study. The dataset comprises source news articles on various topics, reference summaries, machine-generated summaries from different models, and human-evaluated quality scores (from 1.0 to 5.0) across four dimensions: Coherence, Fluency, Consistency, and Relevance, which are used as ground-truth scores in our study. According to prior studies \cite{zhu2020, ideate}, generated texts commonly possess two key features: internal quality and external alignment. Internal quality refers to grammatical accuracy, readability, and structure, focusing on the intrinsic characteristics of the text. In contrast, external alignment evaluates adherence to prompts and factual accuracy, assessing how well the summary reflects and aligns with the source text. Based on the rubric from the original paper \cite{summeval}, we categorize Coherence and Fluency under internal quality, and Consistency and Relevance under external alignment. The detailed definition of each dimension is as follows \cite{summeval}:

\begin{figure*}[t]
    \centering
    \includegraphics[height=0.51\linewidth, width=\linewidth]{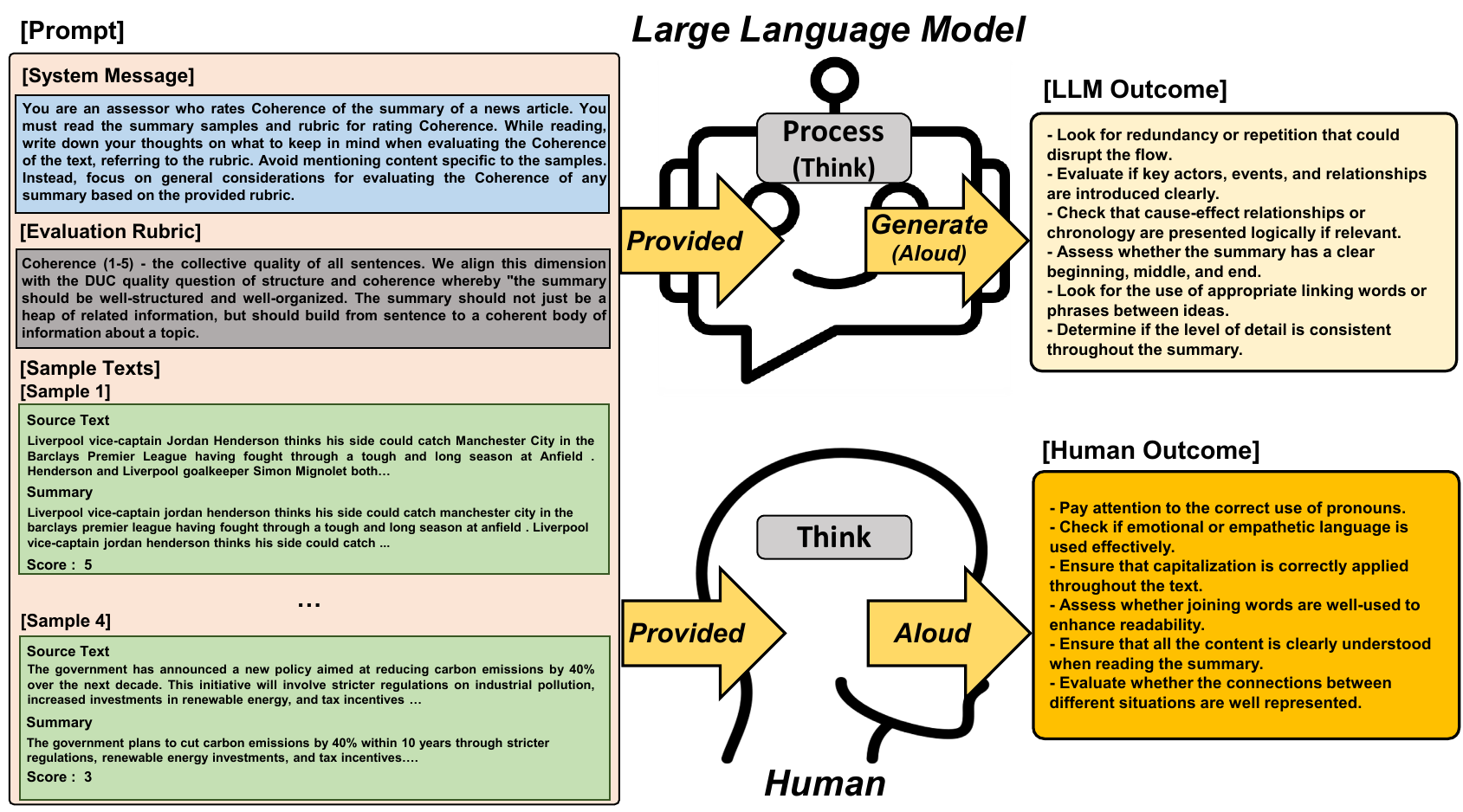}
    \caption{Text Attribute Generation Process: Human experts and LLMs are provided with a system message, a target dimension rubric, sample texts for the dimension, and their corresponding scores. They then review the materials, offering their ideas regarding text attributes that should be considered when rating the dimension based on their knowledge. Technically speaking, humans verbalize text attributes aloud through the process of thinking, whereas large language models (LLMs) generate text attributes after processing input prompt.}
    \label{fig3}
\end{figure*}

\begin{table*}[t]
\centering
\caption{Information of human experts who participated in the TA process.}
\begin{tabular}{llll}
\hline
Expert & Occupation & Related Experience & Language Use (English) \\ \hline
1 & Ph.D Student & Researching on digital and AI journalism & Fluent (non-native) \\
2 & Journalist & News article generation with generative models & Fluent (non-native) \\
3 & University Lecturer & Teaching English writing in a university & Fluent (native) \\
4 & University Lecturer & Teaching English writing in a university & Fluent (native) \\ \hline
\end{tabular}%
\label{table_sum_ta_info}
\end{table*}

\begin{itemize}
    \item \textbf{Internal Quality}
    \begin{itemize}
        \item \textbf{Coherence}: The collective quality of all sentences. We align this dimension with the document understanding conference (DUC) \cite{dang} quality question of structure and coherence whereby "the summary should be well-structured and well-organized. The summary should not just be a heap of related information, but should build from sentence to a coherent body of information about a topic."
        \item \textbf{Fluency}: The quality of individual sentences. Drawing again from the DUC \cite{dang} quality guidlines, sentences in the summary "should have no formatting problems, capitalization errors or obviously ungrammatical sentences (e.g., fragments, missing components) that make the text difficult to read."
    \end{itemize}
    \item \textbf{External Alignment}
    \begin{itemize}
        \item \textbf{Consistency}: The factual alignment between the summary and the summarized source. A factually consistent summary contains only statements that are entailed by the source document. Annotators were also asked to penalize summaries that contained hallucinated facts.
        \item \textbf{Relevance}: Selection of important content from the source. The summary should include only important information from the source document. Annotators were instructed to penalize summaries which contained redundancies and excess information.
    \end{itemize}
\end{itemize}

\subsection{Attribute Collection through Think Aloud} \label{method_think_aloud}

In this stage, we collected the text attributes to construct checklists of each dimension, utilizing the TA process, which enhances divergent thinking \cite{ericsson2, lewis}. The text attributes indicate the key considerations for text evaluation, which are used to design assessment checklists. During the TA process, four human participants verbalized their thoughts while conducting experimental tasks \cite{ericsson1}. Likewise, four different LLMs articulated the text attributes needed for rating the text quality of each dimension, as shown in Figure \ref{fig2} (A). To minimize idiosyncratic biases arising from individual members of each participant group, whether humans or LLMs, we included four members in each group. As illustrated in Figure \ref{fig3}, humans engaged in the TA process by reading the definitions of each dimension alongside four sample summaries per dimension, and the LLMs generated text attributes using the same materials provided to the humans. As a result, the human participants and LLMs generated fine-grained attributes by analyzing the detailed features of each dimension as manifested in the example summaries. Between the methods of CTA and RTA \cite{ericsson2}, which are two main methods of TA, the current framework used CTA, as it is straightforward to implement \cite{concurrent} and free from post-task retrospection biases \cite{van, mcdonald}, with minimal impact on participants' behavior \cite{hertzum}.

\subsubsection{TA by human experts} The human experts consisted of two journalism and two linguistics specialists, chosen based on the nature of the news article summary dataset. We assumed that journalism experts would focus on aspects related to news content while linguistics experts would concentrate on the language features of the texts. Their details are provided in Table \ref{table_sum_ta_info}. The experts conducted the process in real-time via online video calls, addressing each dimension—Coherence, Fluency, Consistency, and Relevance—separately. Specifically, we collected participants' thoughts on each dimension through the following process. First, participants were assigned an assessor persona and provided with four news articles and summaries, actual scores corresponding to different score ranges, and the rubric of each dimension. Then, participants reviewed the news summaries corresponding to each score range and verbalized their reasoning for why they believe the news articles received the given scores. Finally, the participants' verbalizations were recorded and transcribed, and the identified text attributes were compiled.

\begin{figure*}[t]
    \centering
    \includegraphics[height=0.2\linewidth, width=0.7\linewidth]{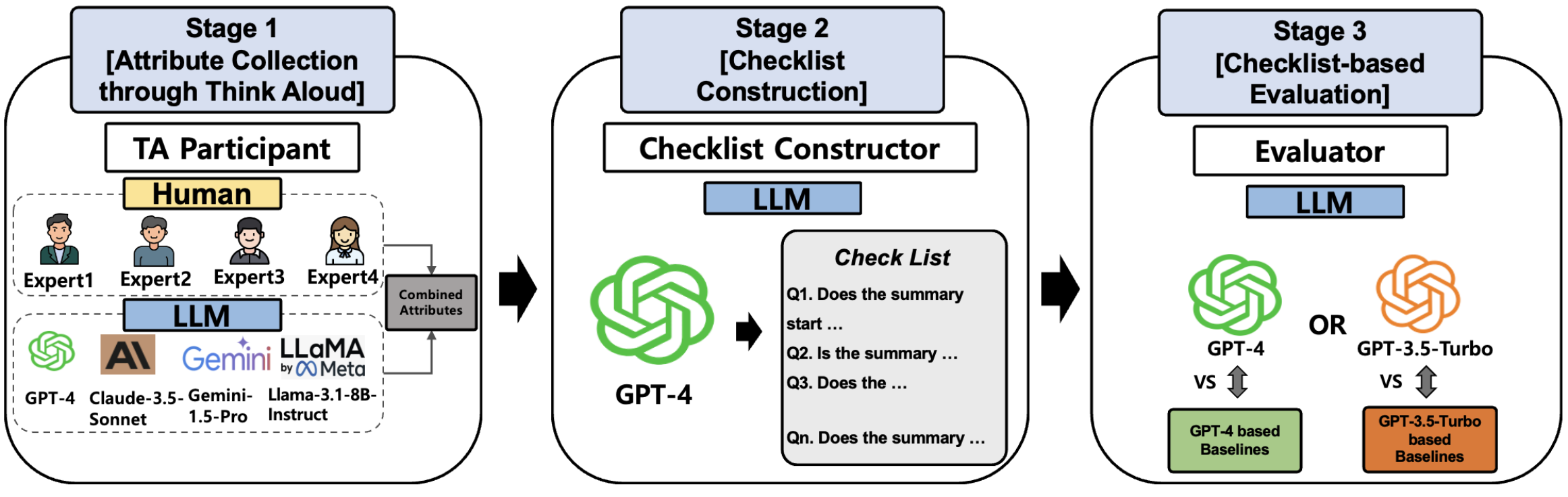}
    \caption{The Overall Process of Text Evaluation: \textbf{Stage 1}: Four humans and four LLMs participate in think-aloud-based text attribute generation, respectively. \textbf{Stage 2}: GPT-4 is solely utilized to construct the checklists. \textbf{Stage 3}: Either GPT-4 or GPT-3.5-Turbo is used to evaluate the quality of the summaries, which are presented with the checklists to guide the evaluation process.}
    \label{stage_fig}
\end{figure*}


\begin{table}[t]
\centering
\caption{System message and predefined template for LLMs' TA}
\resizebox{\linewidth}{!}{ 
\begin{tabular}{p{\linewidth}}
\hline
\multicolumn{1}{c}{\textbf{System Message}}      \\ \hline
You are an assessor who rates \{Dimension\} of the summary of a news article. \\ You must read the summary samples and rubric for rating \{Dimension\}. \\ While reading, write down your thoughts on what to keep in mind when evaluating the \{Dimension\} of the text, referring to the rubric. \\ Avoid mentioning contents specific to the samples. \\ Instead, focus on general considerations for evaluating the \{Dimension\} of any summary based on the provided rubric. \\ \hline
\multicolumn{1}{c}{\textbf{Predefined Template}} \\ \hline
Rubric of \{Dimension\}:\\\{Rubric\}\\ \textless{}Sample 1\textgreater\\ News Article: \{Source Text\}\\ Summary: \{Summary\}\\ Score: \{Score\}\\ \textless{}End of Sample 1\textgreater\\ ...\\ \textless{}Sample 4\textgreater\\ News Article: \{Source Text\}\\ Summary: \{Summary\}\\ Score: \{Score\}\\ \textless{}End of Sample 4\textgreater\\ Please provide each consideration in a JSON format with the keys of consideration index and the values of consideration. \\ Do not name each consideration.\\ For example, \{1: "consideration 1", 2: "consideration 2", 3: "consideration 3", 4: "consideration 4", 5: "consideration 5", ...\} \\ \hline
\end{tabular}%
}
\label{llm_ta_prompt}
\end{table}

\subsubsection{TA by LLMs} The LLMs included four different types of representative agent: GPT-4 \cite{gpt4}, Llama-3.1-8B-Instruct \footnote{https://llama.meta.com/}, Claude-3.5-Sonnet \footnote{https://www.anthropic.com/news/claude-3-5-sonnet}, and Gemini-1.5-Pro \cite{gemini}. The LLMs simulated a human-like TA process using prompt templates based on the TA method as shown in Table \ref{llm_ta_prompt}. In reality, unlike humans, LLMs generate text attributes not through TA but through a process of synthesizing responses to prompts based on their pre-trained background knowledge and text-generation algorithms. However, for ease of exposition, this process is referred to as TA in our paper. The same prompt templates were used across the four mentioned LLMs for attribute generation. For this TA activity, we constructed a prompt template per dimension using the identical materials given to humans during the TA procedure: four news articles and summaries, corresponding scores, and dimension definitions. During the TA process, we assigned the role of an assessor to the LLMs using prompt templates, mirroring the human TA process, where we specifically assigned human participants the assessor role.

\subsection{Checklist Construction} \label{method_checklist_construction}
In this stage, we merged the attributes collected from the previous TA process (Section \ref{method_think_aloud}) of both humans and LLMs. Then, we built the checklists using the combined text attributes. The checklist construction process consists of three steps: 1) Component Extraction, 2) Attribute Clustering, and 3) Question Generation. Each step can utilize any off-the-shelf LLM, such as GPT series, and in our study, we used GPT-4 identically as shown in Figure \ref{stage_fig}. The integration of attributes from both sources (humans and LLMs) was expected to create a synergistic effect, where the strengths of human expertise and LLMs complement each other, leading to a more refined and effective checklist.

\begin{table}[t]
\centering
\caption{System message and predefined template for component extraction}
\resizebox{\linewidth}{!}{ 
\begin{tabular}{p{\linewidth}}
\hline
\multicolumn{1}{c}{\textbf{System Message}} \\ \hline
Given a set of attributes, extract the components continually mentioned in the attributes based on the following conditions. \\ 
1. The components should be related to \{Dimension\} defined below.\\ 
2. Remove any components that are ambiguous or redundant.\\ 
3. The number of the components should not be more than 5.\\ 
Rubric of \{Dimension\}: \{Rubric\} \\ \hline
\multicolumn{1}{c}{\textbf{Predefined Template}} \\ \hline
Attributes: \{Attributes\}\\ 
Please, provide your answer in a List format with the list of components.\\ 
For example, {[}"component 1", "component 2", "component 3", ...{]} \\ \hline
\end{tabular}
}
\label{component_prompt}
\end{table}

\subsubsection{Component extraction}
The component extraction process identified key themes related to a rating dimension from the text attributes generated in the prior TA process. The components represent key sub-concepts within a dimension, specifying the primary areas of assessment. For instance, a component related to Coherence may involve attributes such as "Structure", "Logical Flow" and "Tone style." This step narrows down the focus areas, laying the groundwork for generating key questions tied to these components. Using an LLM as an extractor, principle components were extracted from the merged attributes by prioritizing the most pertinent attributes from both human experts and the LLMs. To avoid noise and redundancy, we prompted the extractor LLM to eliminate ambiguous or repetitive components from the combined attributes. The number of components was limited to five or fewer to prevent excessive subdivision, which could lead to overlapping meanings and create noise. These components then guided the subsequent steps in checklist development. Table \ref{component_prompt} represents the prompt template leveraged for extracting components.

\subsubsection{Attributes clustering}
After extracting the components from the combined group of attributes, an LLM clustered each attribute under the appropriate component to identify specific associated attributes. This clustering divided the scoring dimension into segmented sections, leveraging the ideas of both humans and LLMs. Each section focused on a specific aspect of the dimension, enabling more targeted evaluation. For example, under the "Logical Flow" component, attributes such as "Evaluate whether the connections between different situations are well represented," "Check if the descriptions of the same subject are well-linked," and "Check if the sentences are related to each other" were clustered. Table \ref{attributes_prompt} shows the prompt template with regard to attributes clustering.

\begin{table}[t]
\centering
\caption{System message and predefined template for attributes clustering}
\label{attributes_prompt}
\resizebox{\linewidth}{!}{ 
\begin{tabular}{p{\linewidth}}
\hline
\multicolumn{1}{c}{\textbf{System Message}} \\ \hline
Given a set of attributes and components, group the attributes based on the components.\\ You can remove attributes that are redundant or confusing. \\ \hline
\multicolumn{1}{c}{\textbf{Predefined Template}} \\ \hline
Components: \{Components\}\\ Attributes: \{Attributes\}\\ Please, provide your answer in a JSON format with the keys of components and the values of the list of attributes.\\For example, \{"component 1": {[}"attribute 1", "attribute 4"{]}, "component 2": {[}"attribute 2", "attribute 3"{]}, ...\} \\ \hline
\end{tabular}%
}
\end{table}

\begin{table}[t]
\centering
\caption{System message and predefined template for question generation}
\label{question_prompt}
\resizebox{\linewidth}{!}{ 
\begin{tabular}{p{\linewidth}}
\hline
\multicolumn{1}{c}{{\textbf{System Message}}}      \\ \hline
You are a key question generator for evaluating \{Dimension\}.\\ You have been provided with components along with their associated attributes. Your task is to:\\ 1. Identify and extract common features that appear across the attributes of each component.\\ 2. Based on these common features, construct a set of Yes-No questions related to each key component. \\ These questions should be designed to effectively evaluate the accuracy and relevance of a summary of a news article.\\ Ensure that your response is well-organized, clear, and meets the highest standards of accuracy and reliability. \\ \hline
\multicolumn{1}{c}{{\textbf{Predefined Template}}} \\ \hline
The rubric of \{Dimension\} and the set of components and their attributes corresponding to \{Dimension\} are provided below.\\ Use this information to generate Yes-No questions of the key components.\\ Each component and question must satisfy the following conditions:\\ 1. Each component should have one corresponding question.\\ 2. Each question must be answerable with "Yes" or "No".\\ 3. Each question must incorporate concepts from the key component.\\ 4. Each question should minimize the subjectivity of the rater’s judgment.\\ 5. Formulate questions so that a "Yes" answer is a positive response.\\ \# Rubric\\ \{Rubric of a dimension\}\\ \# Components and attributes\\ \{Components and attributes\}\\ Please provide your answer in a JSON format with the keys of components and the values of questions. \\For example, \{"component 1": "question", "component 2": "question", "component 3": "question", ...\}\\ The word "component" should not appear in the keys of the answer format. \\ \hline
\end{tabular}%
}
\end{table}

\subsubsection{Question generation}
In this stage, the clustered attributes from the prior step were transformed into actionable "Yes-No" questions aimed at probing the identified attributes. First, the LLM generated a key question for each component based on the corresponding attributes. For example, if there were five components, the LLM generated five key questions grounded on each component. Subsequently, the agent broke down the key questions into detailed
sub-questions to more precisely identify the relevant features within a summary. These questions form the core of the checklist, guiding an LLM evaluator in assessing the summaries.

The generated questions underwent a review by a validator LLM to combine any redundant questions and drop irrelevant questions, ensuring a checklist is both reliable and practical for final use in rating texts. The questions should be answerable with "Yes" or "No" containing concepts of a rating dimension as well. After the validation process, the final questions composed the checklist that was used for text evaluation. The prompt templates utilized in this stage are shown in Tables \ref{question_prompt}, \ref{sub_question_prompt}, and \ref{question_validation_prompt}. They were used at different phases of the stage. Table \ref{question_prompt} details the template for generating questions, Table \ref{sub_question_prompt} outlines the template for breaking down these questions into sub-questions, and Table \ref{question_validation_prompt} provides the template used for question validation. They were inspired by the prompts introduced in \cite{checkeval}. The generated checklist for "Coherence" is shown in Table \ref{coherence_checklist} (see Appendix \ref{appedix_checklist} for the checklist of other dimensions).

\begin{table}[t]
\centering
\caption{System message and predefined template for sub-question generation}
\label{sub_question_prompt}
\resizebox{\linewidth}{!}{ 
\begin{tabular}{p{\linewidth}}
\hline
\multicolumn{1}{c}{{\textbf{System Message}}} \\ \hline
You are a sub-question generator. You must construct Yes-No sub-questions for each component. \\ \hline
\multicolumn{1}{c}{{\textbf{Predefined Template}}} \\ \hline
In this task, you need to create a question to evaluate the \{Dimension\} of the summary of the original document. \\ The rubric of \{Dimension\} and the questions corresponding to the key component of \{Dimension\} are provided below. \\ Use them to generate sub-questions for each key question.\\ Each sub-question must satisfy the following conditions:\\ 1. Each question must be answerable with "Yes" or "No".\\ 2. Each sub-question must incorporate concepts from the key component.\\ 3. Each question should minimize the subjectivity of the rater’s judgment. \\ 4. The semantic redundancy between sub-questions should be minimized. \\ 5. Formulate sub-questions so that a "Yes" answer is a positive response.\\ 6. Each sub-question must focus on asking about the presence of positive aspects rather than the absence of negative ones. \\ For example, ask whether a summary avoids negative elements rather than contains them.\\ \# Rubric\\ \{Rubric of a dimension\}\\ \# Components and corresponding questions\\ \{Components and questions\}\\ Please provide your answer in a JSON format with the keys of components and the values of the list of sub-questions.\\ For example, \{"component 1": {[}"sub question 1", "sub question 2",...{]} , "component 2": {[}"sub question 1", "sub question 2", "sub question 3", ...{]},...\} \\ \hline
\end{tabular}%
}
\end{table}

\begin{table}[t]
\centering
\caption{System message and predefined template for question validation}
\label{question_validation_prompt}
\resizebox{\linewidth}{!}{ 
\begin{tabular}{p{\linewidth}}
\hline
\multicolumn{1}{c}{{\textbf{System Message}}} \\ \hline
You are a sub-question evaluator. \\ \hline
\multicolumn{1}{c}{{\textbf{Predefined Template}}} \\ \hline
In this task, you need to examine each sub-question for evaluating the \{Dimension\} of the summary of the original document. \\ The rubric of \{Dimension\} is provided below. \\ \# Rubric\\ \{Rubric of a dimension\}\\ Each sub-question must satisfy the following conditions:\\ 1. Each question must be answerable with "Yes" or "No".\\ 2. Each question must contain concepts of the \{Dimension\}.\\ 3. Each question should minimize the subjectivity of the rater’s judgment. \\ 4. The semantic redundancy between sub-questions should be minimized. \\ 5. Formulate questions so that a "Yes" answer is a positive answer.\\ 6. Please combine any questions that ask about similar characteristics into a single question.\\ 7. Each sub-question must focus on asking about the presence of positive aspects rather than the absence of negative ones. \\ For example, ask whether a summary avoids negative elements rather than contains them.\\ \# Sub-questions\\ \{Sub-questions\}\\ Edit or exclude the problematic questions if it is necessary. Please provide your ultimate answer in a List format.\\ For example, {[}"sub question 1", "sub question 2", "sub question 3", ...{]} \\ \hline
\end{tabular}%
}
\end{table}

\begin{table}[t]
\centering
\caption{Checklist of Coherence dimension generated via the checklist construction process}
\label{coherence_checklist}
\begin{tabularx}{\linewidth}{p{0.1\linewidth}X}
\hline
\multicolumn{1}{c}{\textbf{Dimension}} & \multicolumn{1}{c}{\textbf{Questions}} \\ \hline
Coherence & 
    - Does the summary present information in a logical sequence? \newline
    - Are there smooth transitions between sentences and topics in the summary? \newline
    - Does the summary avoid redundancy or repetition that could disrupt the flow of information? \newline
    - Does each sentence in the summary contribute to the overall structure or flow of ideas? \newline
    - Are pronouns and references used in the summary clear and consistent? \newline
    - Are key actors, events, and relationships introduced clearly in the summary? \newline
    - Is the level of detail maintained consistently throughout the summary? \newline
    - Does the summary avoid sudden shifts in the level of detail? \newline
    - Does the summary provide a balanced amount of detail for all key points? \\ \hline
\end{tabularx}
\end{table}

\subsection{Checklist-based Evaluation} \label{method_evaluation}
As mentioned before, we adopt checklist-based evaluation approach using the checklist created following aforementioned procedures. Therefore, an LLM evaluator is used to directly respond to each checklist question in a binary manner (Yes or No) during this stage. Any LLM can be used at this stage. In our case, as shown in Figure \ref{stage_fig}, we use GPT-3.5-Turbo as the evaluator when comparing our performance with the baselines that use GPT-3.5-Turbo, and we use GPT-4 as the evaluator when comparing with baselines that leverage GPT-4. The number of positive responses (‘Yes’ answers) is then tallied. The proportion of positive responses from the evaluator to the total number of questions is scaled from 1.0 to 5.0 as the final score, to be aligned with the score range of the ground truth scores in the dataset. This method assumes that the ratio of positive responses reflects overall text quality, with each question targeting a specific evaluation component. The scaled scores of each summary are compared with the ground truth scores assigned by humans, which are provided in the dataset.

\section{Experiment}
\label{experiment}
\begin{figure}[t]
    \centering
    \includegraphics[width=\linewidth]{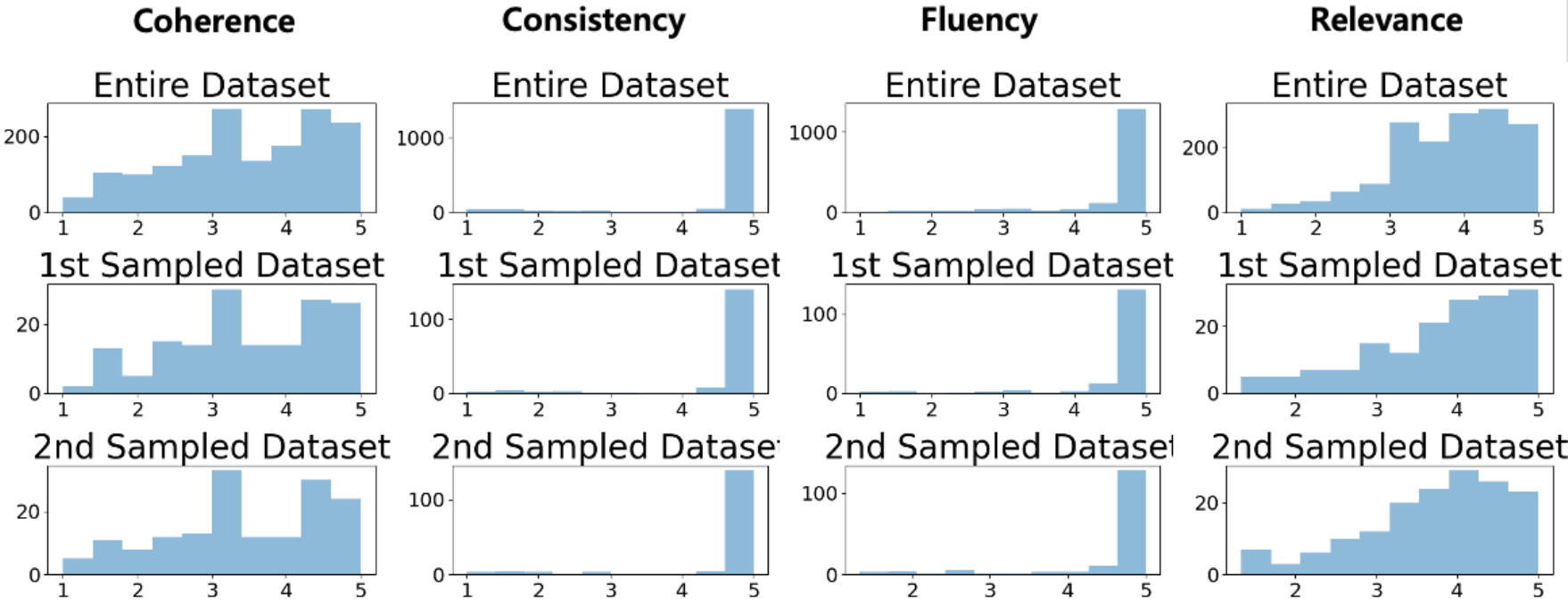}
    \caption{Comparison of Distributions across Four Dimensions in SummEval Dataset: \textbf{Entire Dataset} refers to the distribution of the entire dataset with respect to each dimension, while \textbf{1st Sampled Dataset} and \textbf{2nd Sampled Dataset} represent the distributions of subsets obtained by sampling 10\% of the entire dataset.}
    \label{fig_sample_distribution}
\end{figure}

The experiment consists of three phases aligned with the research questions. First, we compare our method, InteractEval, with baseline text evaluation methods. Next, we assess the effectiveness of the TA approach in creating checklists for text evaluation. Finally, we explore the impact of combining TA outcomes from humans and LLMs, comparing it with other TA conditions: Single-LLM-TA, Single-Human-TA, Multiple-LLMs-TA, and Multiple-Humans-TA. We employ both quantitative metrics, such as correlation with ground-truth scores, and qualitative analysis, focusing on the generated attributes. The code is available at Github repository \footnote{\url{https://github.com/BBeeChu/InteractEval.git}}.

\subsection{Dataset}
We mainly evaluated our InteractEval framework on SummEval dataset \cite{summeval}, a widely recognized benchmark in the text generation task particularly oriented to news article summarization. SummEval contains 1,600 data instances providing source texts, reference texts, and model-generated summaries, evaluated on four dimensions: Coherence, Fluency, Consistency, and Relevance. Each dimension was independently scored by three human annotators using a scale ranging from 1.0 to 5.0, with the final score for each dimension determined by averaging their individual ratings. Additionally, to examine the generalizability of InteractEval's performance on a different text evaluation task, we utilized the ELLIPSE dataset \cite{ellipse}, comprising 6,482 English essays written by non-native ESL (English as a second language) learners. Each essay was scored on seven dimensions: Overall, Cohesion, Conventions, Grammar, Phraseology, Syntax, and Vocabulary. These dimensions were respectively evaluated by two human assessors on a scale of 1.0 to 5.0, and the final score was calculated as the average of their ratings.

\subsection{TA Condition Description}
We examine the effectiveness of the TA process in generating attributes for the checklist used
in text evaluation. We compare rating performances and qualitatively analyze the text attributes across five different TA conditions: Single-LLM-TA (involving a single LLM articulating text attributes, hereafter referred to as SL-TA), Single-Human-TA (involving one human expert verbalizing
one's thought process, SH-TA), Multiple-LLMs-TA (involving several
LLMs articulating text attributes independently, ML-TA), Multiple-Humans-TA (involving multiple human experts verbalizing their thought process independently, MH-TA), and Combination-TA (combining human experts' and LLMs' TA outcomes, Comb-TA). 

\subsection{Quantitative Analysis}
\subsubsection{Measuring correlation}
We measure the performance of InteractEval using \textbf{sample-level correlation}, a method widely adopted in prior LLM-based studies \cite{checkeval, gptscore, chen2023, ahia2023all, wang2024}. This approach helps reduce evaluation costs while maintaining effective performance assessment. Although prior studies have noted the high cost problem of using GPT-series models with large datasets \cite{chen2023frugalgpt, ong2024routellm}, we believe that instance-level evaluation in real-world applications can alleviate this concern. Therefore, we continue using GPT-series models, which are a leading choice for text evaluation. This method computes correlation for each individual sample based on summaries from multiple source texts and then averages these correlations across all samples \cite{checkeval}. For each source text \( x_i \), where \( i \in \{1,2,\dotsc,I\} \), there are \( M \) system-generated summaries \( y_{i,m} \), with \( m \in \{1,2,\dotsc,M\} \). Let \( f_{\text{auto}} \) denote the automatic evaluation metric (e.g., BLEU) and \( f_{\text{human}} \) represent the human evaluation scores. The sample-level correlation \( C^{\text{sample}} \) for each dimension is defined as:

\begin{equation}
    \label{eq2}
    \begin{aligned}
    C^{\text{sample}}_{(f_\text{auto}, f_\text{human})} = \frac{1}{I}\sum_{i=1}^{I}g\Big(&[f_{\text{auto}}(y_{i,1}), \dotsc, f_{\text{auto}}(y_{i,M})], \\
    & [f_{\text{human}}(y_{i,1}), \dotsc, f_{\text{human}}(y_{i,M})]\Big),
    \end{aligned}
\end{equation}
where \( g \) is a correlation coefficient such as Spearman's Rho ($\rho$) or Kendall's Tau ($\tau$). To ensure the representativeness of sample-level correlations, we sampled 10\% data instances from the original dataset while preserving the identical distribution of human annotations (ground truth scores) across each dimension as found in the original one. Figure \ref{fig_sample_distribution} demonstrates that the score distributions of the entire SummEval dataset and the sampled subsets for each dimension are nearly identical, confirming their representativeness (see Appendix \ref{distribution_of_ellipse} for the score distributions in ELLIPSE dataset). Additionally, we evaluated our method's performance against baselines in two trials to assure the robustness of our framework.


\subsubsection{Measuring difference}
In this study, the \textbf{Fisher's transformation} \cite{fisher} is employed to normalize and compare the correlation coefficients of the five separate TA conditions. Converting the correlation values into z-scores allows more accurate comparison and analysis of correlations, particularly in hypothesis testing and constructing confidence intervals.

\subsubsection{Measuring similarity}
We conduct text analysis to examine the similarity of generated attributes within the same dimension. This analysis evaluates whether each TA condition produces diverse attributes from both a lexical and a semantic perspective. For lexical similarity, we use ROUGE \cite{rouge} and Jaccard similarity \cite{jaccard, niwattanakul}, which focus on overlapping words or sequences without considering meaning or context. For semantic similarity, we use cosine similarity, which measures semantic closeness rather than exact word matching.

\subsubsection{Baselines}
The following baselines, including widely used metrics (e.g. ROUGE-L \cite{rouge}, BLEU \cite{bleu}, \& METEOR \cite{meteor}), are compared with our InteractEval.
\begin{itemize}
    \item BERTScore \cite{bertscore}: measures text similarity using semantic embeddings generated by BERT \cite{bert}.
    \item MoverScore \cite{moverscore}: improves BERTScore by integrating soft alignments and innovative aggregation methods for a more accurate similarity assessment.
    \item BARTScore \cite{bartscore}: acts as a unified evaluator by utilizing the pretrained BART model \cite{bart}, which relies on the average likelihood of the model's output.
    \item UniEval \cite{unieval}: functions as a unified multi-dimensional evaluator capable of evaluating different dimensions of text generation tasks.
    \item G-Eval \cite{geval}: employs LLMs for evaluating text quality, utilizing a chain-of-thoughts method and a form-filling framework to rate the quality of texts.
    \item CheckEval \cite{checkeval}: uses checklists to assess text quality by breaking down evaluating dimensions and generating questions for each. Then, it gathers LLM responses to the checklist to calculate rating scores.
\end{itemize}

\subsection{Qualitative analysis}
We conducted a focused group interview (FGI) \cite{morgan, dilshad} with four human experts (E1, E2, E3, and E4) to qualitatively analyze the attributes produced by the humans (A) and LLMs (B) during the TA process. The experts compared both sets of attributes by responding to open-ended questions for each dimension (see Appendix \ref{open_ended_question} for the details of questions). These questions allowed us to gain deeper insights into the attributes and encouraged detailed expert feedback. Three of the experts had participated in the TA process, while one had not. Initially, we planned for all four TA participants to conduct the qualitative analysis, assuming their familiarity would lead to more refined insights. However, one expert with linguistics expertise withdrew for a personal reason. Hence, we invited another linguist, who had been also researching large language models, to participate. After collecting the experts' responses through email, we identified keywords mentioned by the majority (at least three experts). We then filtered out answers containing the keywords. The results will be further discussed in section \ref{humans_strengths} and \ref{llms_strengths}.

\section{Results}
\label{results}
\subsection{Analysis of text evaluation performance (\textbf{RQ}\ref{rq1})}
To examine the performance of InteractEval, we have conducted a series of experiments including a comparison against baseline text evaluation methods, examination of changes in model performance based on the number of components, and generalization assessment of model performance using a different dataset.

\subsubsection{\textbf{Performance comparison with baselines}}
We conducted a comparison against baseline evaluation methods to evaluate the performance of InteractEval. Building upon prior studies \cite{gptscore, summeval, checkeval, geval}, we analyzed the correlation between InteractEval and ground-truth scores using Spearman's Rho, Kendall's Tau coefficients, and the difference between predicted scores and human-labeled scores using mean absolute errors (MAE). For a fair comparison, we used the performance data of the seven non-LLM-based traditional methods (see Table \ref{table2}), G-Eval (GPT-4), and CheckEval (GPT-3.5 Turbo and GPT-4), as reported in \cite{checkeval}, given that their experiments were conducted using the same experimental settings (sample-level correlation). Additionally, we reproduced the correlation results of G-Eval (GPT-3.5-Turbo)\footnote{https://github.com/nlpyang/geval.git} and the MAE results of the entire baselines at the sample level using the source code provided by the original authors (where available) and following the implementation details described in their papers.

\begin{table*}[t]
\centering
\caption{Sample-level performance of different dimensions on SummEval benchmark dataset measured by Spearman's Rho ($\rho$), Kendall's Tau ($\tau$), and Mean Absolute Error (\textit{m}). The best results are highlighted in \textbf{bold}, and the second-best results are \underline{underlined}.}
\label{table2}
\resizebox{\textwidth}{!}{%
\begin{tabular}{cccccccccccccccc}
\hline
\multicolumn{1}{c|}{{}} &
  \multicolumn{6}{c|}{{Internal Quality}} &
  \multicolumn{6}{c|}{{External Alignment}} &
  \multicolumn{3}{c}{{Average}} \\ \cline{2-13}
\multicolumn{1}{c|}{{Models}} &
  \multicolumn{3}{c|}{{Coherence}} &
  \multicolumn{3}{c|}{{Fluency}} &
  \multicolumn{3}{c|}{{Consistency}} &
  \multicolumn{3}{c|}{{Relevance}} &
  {} &
  {} &
  {} \\ \cline{2-16} 
\multicolumn{1}{c|}{{}} &
  {\textit{m} ($\downarrow$)} &
  {$\rho (\uparrow)$} &
  \multicolumn{1}{c|}{{$\tau (\uparrow)$}} &
  {\textit{m} ($\downarrow$)} &
  {$\rho (\uparrow)$} &
  \multicolumn{1}{c|}{{$\tau (\uparrow)$}} &
  {\textit{m} ($\downarrow$)} &
  {$\rho (\uparrow)$} &
  \multicolumn{1}{c|}{{$\tau (\uparrow)$}} &
  {\textit{m} ($\downarrow$)} &
  {$\rho (\uparrow)$} &
  \multicolumn{1}{c|}{{$\tau (\uparrow)$}} &
  {\textit{m} ($\downarrow$)} &
  {$\rho (\uparrow)$} &
  {$\tau (\uparrow)$} \\ \hline
\multicolumn{1}{l}{{Non-LLM based}} &
  \multicolumn{1}{l}{{}} &
  \multicolumn{1}{l}{{}} &
  \multicolumn{1}{l}{{}} &
  \multicolumn{1}{l}{{}} &
  \multicolumn{1}{l}{{}} &
  \multicolumn{1}{l}{{}} &
  \multicolumn{1}{l}{{}} &
  \multicolumn{1}{l}{{}} &
  \multicolumn{1}{l}{{}} &
  \multicolumn{1}{l}{{}} &
  \multicolumn{1}{l}{{}} &
  \multicolumn{1}{l}{{}} &
  \multicolumn{1}{l}{{}} &
  \multicolumn{1}{l}{{}} &
  \multicolumn{1}{l}{{}} \\ \hline
\multicolumn{1}{c|}{{ROUGE-L}} &
  {2.008} &
  {0.172} &
  \multicolumn{1}{c|}{{0.122}} &
  {3.305} &
  {0.240} &
  \multicolumn{1}{c|}{{0.167}} &
  {3.372} &
  {0.317} &
  \multicolumn{1}{c|}{{0.235}} &
  {2.356} &
  {0.420} &
  \multicolumn{1}{c|}{{0.301}} &
  {2.760} &
  {0.287} &
  {0.206} \\
\multicolumn{1}{c|}{{BLEU}} &
  {2.877} &
  {0.028} &
  \multicolumn{1}{c|}{{0.019}} &
  {4.102} &
  {-0.079} &
  \multicolumn{1}{c|}{{-0.057}} &
  {4.214} &
  {0.036} &
  \multicolumn{1}{c|}{{0.030}} &
  {3.215} &
  {0.423} &
  \multicolumn{1}{c|}{{0.308}} &
  {3.602} &
  {0.102} &
  {0.075} \\
\multicolumn{1}{c|}{{METEOR}} &
  {1.632} &
  {0.009} &
  \multicolumn{1}{c|}{{0.008}} &
  {2.679} &
  {0.011} &
  \multicolumn{1}{c|}{{0.008}} &
  {2.813} &
  {0.101} &
  \multicolumn{1}{c|}{{0.074}} &
  {1.847} &
  {0.275} &
  \multicolumn{1}{c|}{{0.182}} &
  {2.243} &
  {0.099} &
  {0.068} \\
\multicolumn{1}{c|}{{BERTScore}} &
  {1.095} &
  {-0.219} &
  \multicolumn{1}{c|}{{-0.151}} &
  {0.715} &
  {-0.156} &
  \multicolumn{1}{c|}{{-0.113}} &
  {\textbf{0.642}} &
  {-0.109} &
  \multicolumn{1}{c|}{{-0.076}} &
  {1.601} &
  {0.462} &
  \multicolumn{1}{c|}{{0.476}} &
  {1.013} &
  {-0.005} &
  {0.034} \\
\multicolumn{1}{c|}{{MOVERScore}} &
  {0.978} &
  {0.340} &
  \multicolumn{1}{c|}{{0.252}} &
  {1.866} &
  {0.345} &
  \multicolumn{1}{c|}{{0.247}} &
  {1.961} &
  {0.384} &
  \multicolumn{1}{c|}{{0.277}} &
  {1.050} &
  {-0.434} &
  \multicolumn{1}{c|}{{-0.315}} &
  {1.464} &
  {0.159} &
  {0.115} \\
\multicolumn{1}{c|}{{BARTScore}} &
  {1.150} &
  {0.475} &
  \multicolumn{1}{c|}{{0.343}} &
  {1.617} &
  {0.395} &
  \multicolumn{1}{c|}{{0.285}} &
  {1.838} &
  {0.457} &
  \multicolumn{1}{c|}{{0.341}} &
  {1.148} &
  {0.602} &
  \multicolumn{1}{c|}{{0.455}} &
  {1.438} &
  {0.482} &
  {0.356} \\
\multicolumn{1}{c|}{{UniEval}} &
  {1.145} &
  {0.567} &
  \multicolumn{1}{c|}{{0.411}} &
  {0.715} &
  {0.605} &
  \multicolumn{1}{c|}{{0.447}} &
  {\textbf{0.642}} &
  {0.613} &
  \multicolumn{1}{c|}{{0.476}} &
  {1.601} &
  {\underline{0.640}} &
  \multicolumn{1}{c|}{{0.476}} &
  {1.026} &
  {0.606} &
  {0.453} \\ \hline
\multicolumn{1}{l}{{LLM based (GPT-3.5-turbo)}} &
  \multicolumn{1}{l}{{}} &
  \multicolumn{1}{l}{{}} &
  \multicolumn{1}{l}{{}} &
  \multicolumn{1}{l}{{}} &
  \multicolumn{1}{l}{{}} &
  \multicolumn{1}{l}{{}} &
  \multicolumn{1}{l}{{}} &
  \multicolumn{1}{l}{{}} &
  \multicolumn{1}{l}{{}} &
  \multicolumn{1}{l}{{}} &
  \multicolumn{1}{l}{{}} &
  \multicolumn{1}{l}{{}} &
  \multicolumn{1}{l}{{}} &
  \multicolumn{1}{l}{{}} &
  \multicolumn{1}{l}{{}} \\ \hline
\multicolumn{1}{c|}{{G-Eval}} &
  {1.351} &
  {0.412} &
  \multicolumn{1}{c|}{{0.393}} &
  {2.750} &
  {0.354} &
  \multicolumn{1}{c|}{{0.333}} &
  {\underline{0.871}} &
  {0.338} &
  \multicolumn{1}{c|}{{0.322}} &
  {0.844} &
  {0.361} &
  \multicolumn{1}{c|}{{0.321}} &
  {1.454} &
  {0.366} &
  {0.342} \\
\multicolumn{1}{c|}{{CheckEval}} &
  {1.135} &
  {0.440} &
  \multicolumn{1}{c|}{{0.347}} &
  {1.078} &
  {0.393} &
  \multicolumn{1}{c|}{{0.327}} &
  {1.694} &
  {0.600} &
  \multicolumn{1}{c|}{{0.515}} &
  {0.851} &
  {0.305} &
  \multicolumn{1}{c|}{{0.252}} &
  {1.190} &
  {0.435} &
  {0.360} \\
\multicolumn{1}{c|}{{InteractEval (1st)}} &
  {1.185} &
  {0.583} &
  \multicolumn{1}{c|}{{0.536}} &
  {1.860} &
  {0.630} &
  \multicolumn{1}{c|}{{0.604}} &
  {1.558} &
  {0.734} &
  \multicolumn{1}{c|}{{0.702}} &
  {0.998} &
  {0.614} &
  \multicolumn{1}{c|}{{0.576}} &
  {1.400} &
  {0.640} &
  {0.605} \\
\multicolumn{1}{c|}{{InteractEval (2nd)}} &
  {1.226} &
  {0.590} &
  \multicolumn{1}{c|}{{0.540}} &
  {1.482} &
  {0.614} &
  \multicolumn{1}{c|}{{0.598}} &
  {1.021} &
  {0.726} &
  \multicolumn{1}{c|}{{0.698}} &
  {0.936} &
  {0.623} &
  \multicolumn{1}{c|}{{0.577}} &
  {1.166} &
  {0.638} &
  {0.603} \\ \hline
\multicolumn{1}{l}{{LLM based (GPT-4)}} &
  \multicolumn{1}{l}{{}} &
  \multicolumn{1}{l}{{}} &
  \multicolumn{1}{l}{{}} &
  \multicolumn{1}{l}{{}} &
  \multicolumn{1}{l}{{}} &
  \multicolumn{1}{l}{{}} &
  \multicolumn{1}{l}{{}} &
  \multicolumn{1}{l}{{}} &
  \multicolumn{1}{l}{{}} &
  \multicolumn{1}{l}{{}} &
  \multicolumn{1}{l}{{}} &
  \multicolumn{1}{l}{{}} &
  \multicolumn{1}{l}{{}} &
  \multicolumn{1}{l}{{}} &
  \multicolumn{1}{l}{{}} \\ \hline
\multicolumn{1}{c|}{{G-Eval}} &
  {1.084} &
  {0.619} &
  \multicolumn{1}{c|}{{0.470}} &
  {2.195} &
  {0.629} &
  \multicolumn{1}{c|}{{0.482}} &
  {1.216} &
  {0.664} &
  \multicolumn{1}{c|}{{0.517}} &
  {1.399} &
  {0.617} &
  \multicolumn{1}{c|}{{0.464}} &
  {1.474} &
  {0.632} &
  {0.483} \\
\multicolumn{1}{c|}{{CheckEval}} &
  {\underline{0.944}} &
  {0.573} &
  \multicolumn{1}{c|}{{0.428}} &
  {0.934} &
  {0.632} &
  \multicolumn{1}{c|}{{0.493}} &
  {1.429} &
  {0.706} &
  \multicolumn{1}{c|}{{0.611}} &
  {\underline{0.684}} &
  {0.570} &
  \multicolumn{1}{c|}{{0.438}} &
  {0.998} &
  {0.620} &
  {0.493} \\
\multicolumn{1}{c|}{{InteractEval (1st)}} &
  {\textbf{0.847}} &
  {\underline{0.649}} &
  \multicolumn{1}{c|}{{\underline{0.603}}} &
  {\textbf{0.356}} &
  {\textbf{0.799}} &
  \multicolumn{1}{c|}{{\textbf{0.769}}} &
  {1.042} &
  {\underline{0.783}} &
  \multicolumn{1}{c|}{{\underline{0.761}}} &
  {0.919} &
  {0.626} &
  \multicolumn{1}{c|}{{\underline{0.587}}} &
  {\textbf{0.791}} &
  {\underline{0.714}} &
  {\underline{0.680}} \\
\multicolumn{1}{c|}{{InteractEval (2nd)}} &
  {1.177} &
  {\textbf{0.660}} &
  \multicolumn{1}{c|}{{\textbf{0.606}}} &
  {\underline{0.639}} &
  {\underline{0.781}} &
  \multicolumn{1}{c|}{{\underline{0.751}}} &
  {1.009} &
  {\textbf{0.816}} &
  \multicolumn{1}{c|}{{\textbf{0.798}}} &
  {\textbf{0.633}} &
  {\textbf{0.642}} &
  \multicolumn{1}{c|}{{\textbf{0.592}}} &
  {\underline{0.865}} &
  {\textbf{0.725}} &
  {\textbf{0.687}} \\ \hline
\end{tabular}%
}
\end{table*}

Table \ref{table2} presents the main results, classifying the four dimensions into two key features by which generated texts are typically assessed: internal quality and external alignment \cite{zhu2020, ideate}. Additionally, the table shows InteractEval's performance in the first and second trials using both GPT-3.5-Turbo and GPT-4, demonstrating stable sample-level correlations. In the both trials, InteractEval with GPT-4 outperforms or is at least comparable to the traditional non-LLM-based metrics, such as ROUGE-L, BLEU, METEOR, BERTScore, MOVERScore, BARTScore, and UniEval, in terms of correlation coefficients. As noted in \cite{checkeval}, the results suggest that using a checklist enables an LLM to evaluate generated text in a manner more aligned with human judgment than traditional automatic metrics. 

When compared to G-Eval and CheckEval with the identical version of GPT, our model demonstrates superior performance. Moreover, even with GPT-3.5-Turbo, our framework outperforms other LLM-based baselines with GPT-4 or shows comparable performance, indicating its cost-saving potential while ensuring reasonable performance \cite{leancontext, ccot, shekhar}. These results highlight the broad capability of our proposed framework, showing that using TA and the checklist constructed through the combined TA results of humans and LLMs lead to accurate and cost-effective evaluations across the four text quality dimensions of text summarization.

In terms of MAE performance, there is no clear consistent trend, although InteractEval with GPT-4 achieves the best performance on average. Specifically, the results show that models with high correlations to human raters do not necessarily produce lower MAE errors. This discrepancy arises from the theoretical differences between MAE, which measures the average absolute difference between predicted and actual scores \cite{mae}, and correlation, which evaluates the strength and direction of a linear relationship between the scores \cite{correlation}. Unlike correlation, MAE does not account for the direction of errors and cannot distinguish between overestimations and underestimations. Taken together, the 
sample-level analysis conducted in relation to human assessment scores as ground-truth indicate that correlation analysis on the basis of Spearman's Rho or Kendall's Tau are more sensitive than the analysis conducted with Mean Absolute Error.

\begin{figure}[t]

    \centering
    \includegraphics[width=\linewidth]{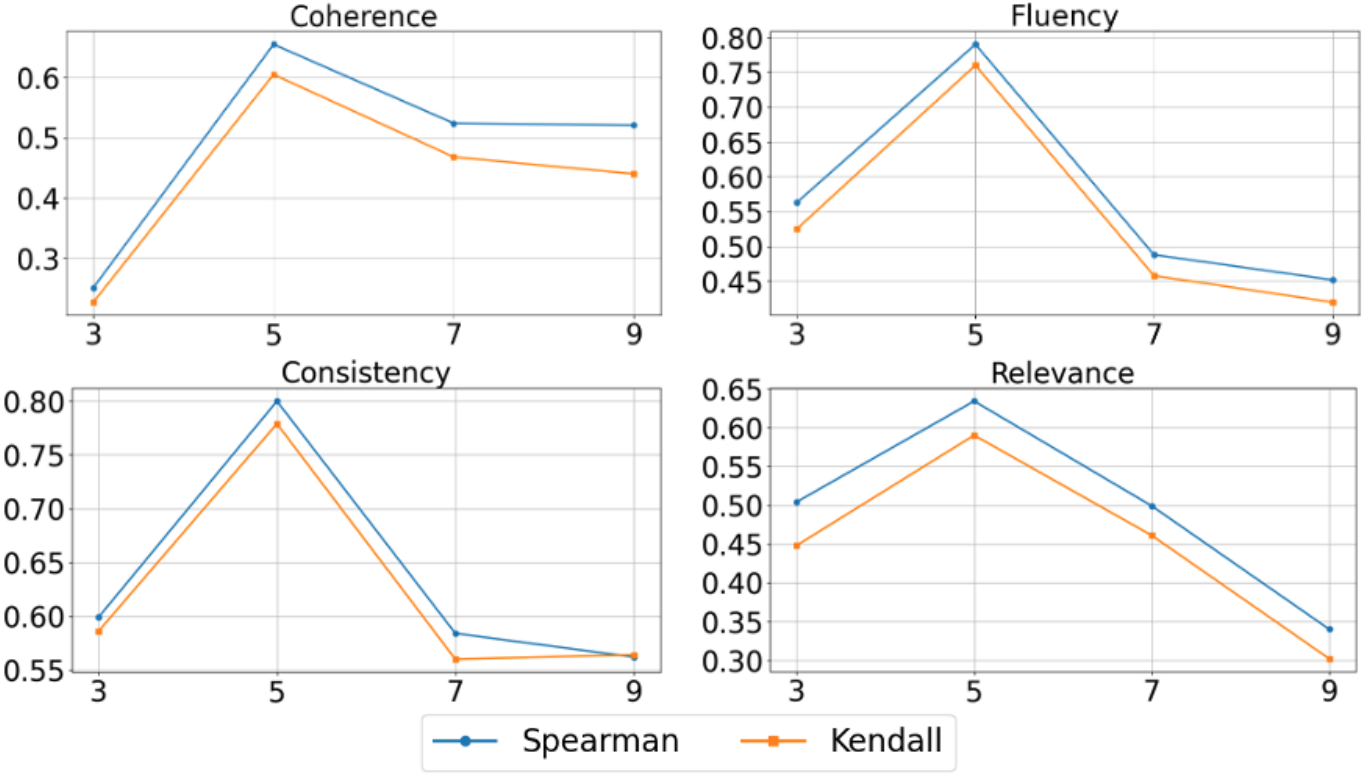}

    \caption{Performance of InteractEval across Different Component Numbers: Two correlation measures (y-axis) are compared across the four evaluation dimensions—Coherence, Fluency, Consistency, and Relevance—by varying the number of components (x-axis) extracted from Comb-TA attributes. The reported correlation values represent the average results from the first and second trials.}
    \label{fig_component}
\end{figure}

\subsubsection{\textbf{Performance comparison with component numbers.}}
As previously mentioned, we prompted an LLM to extract five components from TA attributes before generating questions and constructing checklists. To evaluate the impact of the number of components on InteractEval's performance, we conducted an ablation study, varying the number of components from three to nine, as shown in Figure \ref{fig_component}. The results indicate that our model achieves optimal performance when using checklists constructed with five components derived from TA attributes. We believe these performance differences are reflective of the difficulty of capturing sufficient key sub-concepts related to each evaluation dimension when extremely few components are extracted, and the redundancy that arises when too many components are extracted. Therefore, we empirically confirm through experiments that extracting five components strikes an optimal balance, enabling the identification of the most relevant key sub-concepts for all dimensions and facilitating the creation of the most effective checklists.

\begin{figure*}[t]

    \centering
    \includegraphics[width=\textwidth]{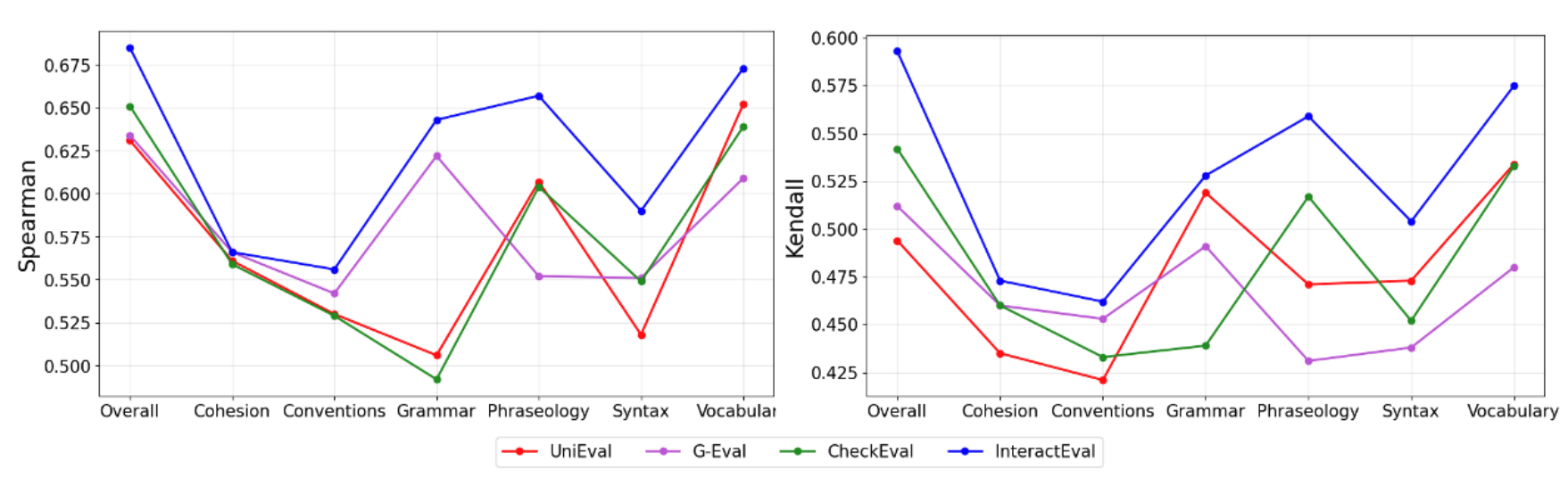}

    \caption{Performance Comparison on ELLIPSE dataset: Sample-level correlations of different dimensions on ELLIPSE benchmark dataset are measured by Spearman's Rho ($\rho$) and Kendall's Tau ($\tau$). The reported results represent the average performance across the first and second trials (see Appendix \ref{ellipse_performance_appendix} for exact values).}
    \label{ellipse_performance}
\end{figure*}

\begin{table*}[t]
\centering
\caption{Sample-level correlations of different dimensions on SummEval benchmark dataset are measured using Spearman's Rho ($\rho$) and Kendall's Tau ($\tau$). Entire models utilized GPT-4 as the evaluator. The model names in parentheses in InteractEval refer to the models used for the TA process. The best results are highlighted in \textbf{bold}, and the second-best results are \underline{underlined}.}
\resizebox{0.8\textwidth}{!}{%
\begin{tabular}{lcccccccccc}
\hline
\multicolumn{1}{l|}{} &
  \multicolumn{4}{c|}{\textbf{Internal Quality}} &
  \multicolumn{4}{c|}{\textbf{External Alignment}} &
  \multicolumn{2}{c}{\textbf{Average}} \\ \cline{2-9}
\multicolumn{1}{l|}{Models} &
  \multicolumn{2}{c|}{\textbf{Coherence}} &
  \multicolumn{2}{c|}{\textbf{Fluency}} &
  \multicolumn{2}{c|}{\textbf{Consistency}} &
  \multicolumn{2}{c|}{\textbf{Relevance}} &
  \textbf{} &
  \textbf{} \\ \cline{2-11} 
\multicolumn{1}{l|}{} &
  $\rho$ &
  \multicolumn{1}{c|}{$\tau$} &
  $\rho$ &
  \multicolumn{1}{c|}{$\tau$} &
  $\rho$ &
  \multicolumn{1}{c|}{$\tau$} &
  $\rho$ &
  \multicolumn{1}{c|}{$\tau$} &
  $\rho$ &
  $\tau$ \\ \hline
  
\multicolumn{1}{l|}{G-Eval} &
  0.619 &
  \multicolumn{1}{c|}{0.47} &
  0.629 &
  \multicolumn{1}{c|}{0.482} &
  0.664 &
  \multicolumn{1}{c|}{0.517} &
  \textbf{0.617} &
  \multicolumn{1}{c|}{0.464} &
  {\underline{0.632}} &
  0.483 \\
\multicolumn{1}{l|}{CheckEval} &
  0.573 &
  \multicolumn{1}{c|}{0.428} &
  0.632 &
  \multicolumn{1}{c|}{0.493} &
  {\underline{0.706}} &
  \multicolumn{1}{c|}{{\underline{0.611}}} &
  {\underline{0.570}} &
  \multicolumn{1}{c|}{0.438} &
  0.620 &
  0.493 \\ \hline
\multicolumn{1}{l|}{InteractEval (GPT-4)} &
  {\underline{0.620}} &
  \multicolumn{1}{c|}{\textbf{0.598}} &
  {\underline{0.633}} &
  \multicolumn{1}{c|}{{\underline{0.600}}} &
  \textbf{0.738} &
  \multicolumn{1}{c|}{\textbf{0.724}} &
  0.550 &
  \multicolumn{1}{c|}{\textbf{0.530}} &
  \textbf{0.635} &
  \textbf{0.613} \\
\multicolumn{1}{l|}{InteractEval (Llama-3.1-8B-Instruct)} &
  0.494 &
  \multicolumn{1}{c|}{0.449} &
  0.570 &
  \multicolumn{1}{c|}{0.554} &
  0.583 &
  \multicolumn{1}{c|}{0.572} &
  0.436 &
  \multicolumn{1}{c|}{0.406} &
  0.521 &
  0.495 \\
\multicolumn{1}{l|}{InteractEval (Gemini-1.5-Pro)} &
  0.453 &
  \multicolumn{1}{c|}{0.413} &
  \textbf{0.658} &
  \multicolumn{1}{c|}{\textbf{0.647}} &
  0.630 &
  \multicolumn{1}{c|}{0.610} &
  0.468 &
  \multicolumn{1}{c|}{0.429} &
  0.552 &
  {\underline{0.525}} \\
\multicolumn{1}{l|}{InteractEval (Claude-3.5-Sonnet)} &
  \textbf{0.632} &
  \multicolumn{1}{c|}{{\underline{0.581}}} &
  0.467 &
  \multicolumn{1}{c|}{0.427} &
  0.522 &
  \multicolumn{1}{c|}{0.462} &
  0.523 &
  \multicolumn{1}{c|}{{\underline{0.470}}} &
  0.536 &
  0.485 \\ \hline
\end{tabular}%
}

\label{table3}
\end{table*}

\subsubsection{\textbf{Evaluation performance on ELLIPSE dataset}}
To provide supplementary evaluation beyond the text summary benchmark dataset and examine the generalizability of InteractEval across different tasks, we employed our method using the ELLIPSE dataset. We followed the same procedures as the SummEval dataset including attribute collection through think aloud, checklist construction, and checklist-based evaluation. Additionally, we conducted the human TA procedure with four linguistic experts, considering that the dataset is related to English language education (see appendix \ref{ellipse_ta_info} for detailed information of the participants). None of the baselines report their performance on the dataset; accordingly, the baseline results are based on our implementation following the experimental settings outlined in their original papers or source codes (if available). Note that, in CheckEval \cite{checkeval}, authors filtered LLM-generated checklists in the final stage; therefore, we followed the same process after obtaining the checklists. As with the SummEval dataset, we sampled 10\% of the original dataset twice and ran InteractEval and the baselines on the two samples (referred to as the first trial and second trial below).

Figure \ref{ellipse_performance} presents the average performance of the top four models across the first and second trials. The results show that InteractEval surpasses all baselines across entire dimensions, demonstrating the broader applicability of InteractEval on text evaluation. However, unlike with the SummEval dataset, our method using GPT-3.5-Turbo does not consistently outperform other baselines, suggesting that the cost-effectiveness of InteractEval depends on the specific dataset used. Because InteractEval produces the best performance outcomes when using GPT-4 as the evaluator, the remaining experiments were all conducted with GPT-4.

\subsection{Analysis of TA effectiveness (\textbf{RQ}\ref{rq2})}

The proposed framework initially collects various attributes from humans and LLMs that should be considered while designing checklists per dimension. Unlike prior research, we utilize TA to encourage LLMs to generate a wider range of text attributes, which form the basis for checklists across different evaluation dimensions. To examine how the TA process affects the evaluation performance of our framework, we compare its performance when using checklists constructed from a single LLM's attributes under the TA process against other LLM-based baselines on the SummEval dataset. As shown in Table \ref{table3}, the performance of InteractEval across the four evaluation dimensions varies depending on the LLM used for the TA process. For each dimension, at least one of the four InteractEval conditions with a single-LLM TA surpasses the baselines, G-Eval \cite{geval} and CheckEval \cite{checkeval}, except for the Spearman's Rho correlation of Relevance. The results show that incorporating TA improves the generation of text attributes by LLMs to some extent. In particular, our method, using single-LLM-TA outcomes with GPT-4, outperforms other LLM-based baselines, except in Spearman's Rho correlation for Relevance. 
For overall average, irrespective of whether the correlation is assessed via Spearman's Rho or Kendall's Tau, InteractEval with single-LLM TA (GPT-4) surpasses all of the baselines, without exception. These results suggest that even with just one LLM, particulary with GPT-4, the TA process leads to more refined checklists, improving evaluation outcomes. By encouraging divergent analysis \cite{ericsson1, ericsson2, lewis}, the TA process enables LLMs to capture more detailed aspects of the dimension rubrics based on sample texts. In the end, the study results show that combining the TA outcomes from the four LLMs complements each other, contributing to removing idiosyncratic differences and overcoming each one's weaknesses, and consequently results in checklists of higher quality.

\begin{figure*}[t]
    \centering
    \includegraphics[width=0.5\textwidth]{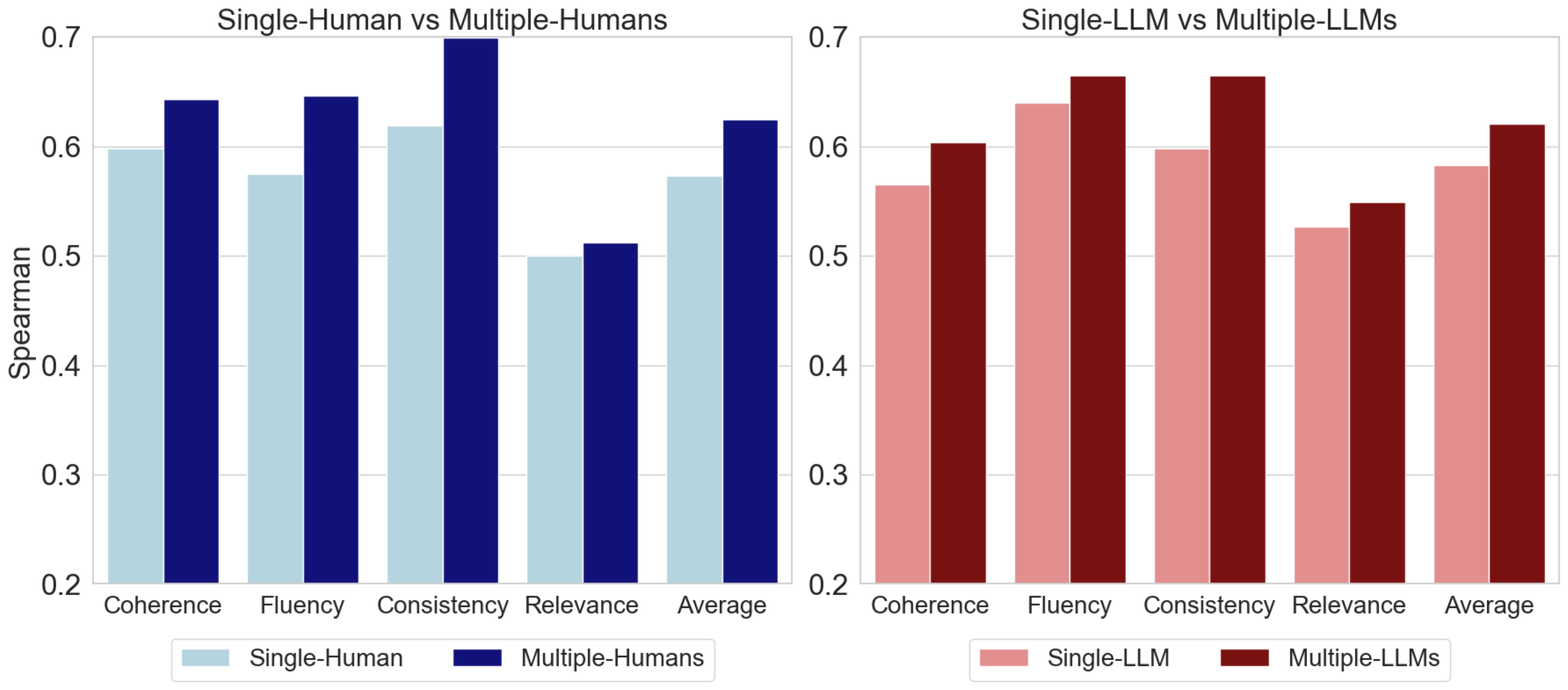}    
    \caption{Performance Comparison Between Single-TA and Multiple-TA Conditions: Evaluation performance in Spearman's Rho ($\rho$) correlations is compared between single-TA and multiple-TA conditions across the four dimensions. Single-Human and Single-LLM represent the average performance results from four human participants and four LLMs, respectively.}
    \label{fig4}
\end{figure*}

\begin{figure*}[t]

    \centering
    \includegraphics[width=\textwidth]{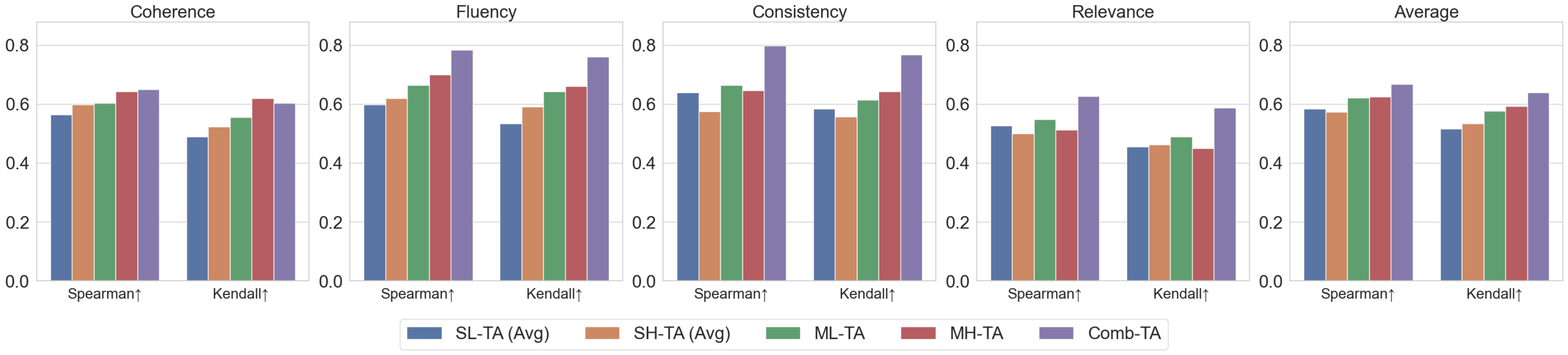}

    \caption{Comparison of two correlation measures (Spearman's Rho ($\rho$) \& Kendall's Tau ($\tau$)) across the four evaluation dimensions - Coherence, Fluency, Consistency, and Relevance. The bars represent the correlations of different TA conditions: \textbf{SL-TA (Avg), SH-TA (Avg), ML-TA, MH-TA, and Comb-TA}. The performance of SL-TA (Avg) and SH-TA (Avg) is based on the average values across the four different checklists created by each LLM and human expert, respectively.}
    \label{fig5}
\end{figure*}

\begin{table*}[t]
\centering
\caption{Comparison of two correlation measures, Spearman's Rho ($\rho$) and Kendall's Tau ($\tau$), over four dimensions-Coherence, Fluency, Consistency, and Relevance. The values are z-scores that indicate the statistical significance of the differences between the method pairs (\textbf{L}: LLM, \textbf{H}: Human, \textbf{S}: Single, \textbf{M}: Multiple, \textbf{C}: Combination), with asterisks (*) denoting the level of significance: * (p-vale \textless 0.05), ** (p-value \textless 0.01), and *** (p-value \textless 0.001).}
\resizebox{0.8\textwidth}{!}{%
\begin{tabular}{llcccccccccc}
\hline
 &
  \multicolumn{1}{l|}{} &
  \multicolumn{4}{c|}{\textbf{Internal Quality}} &
  \multicolumn{4}{c|}{\textbf{External Alignment}} &
  \multicolumn{2}{c}{\textbf{Average}} \\ \cline{3-10}
\multicolumn{2}{c|}{Methods} &
  \multicolumn{2}{c|}{\textbf{Coherence}} &
  \multicolumn{2}{c|}{\textbf{Fluency}} &
  \multicolumn{2}{c|}{\textbf{Consistency}} &
  \multicolumn{2}{c|}{\textbf{Relevance}} &
  \multicolumn{2}{c}{\textbf{}} \\ \cline{3-12} 
 &
  \multicolumn{1}{l|}{} &
  $\rho$ &
  \multicolumn{1}{c|}{$\tau$} &
  $\rho$ &
  \multicolumn{1}{c|}{$\tau$} &
  $\rho$ &
  \multicolumn{1}{c|}{$\tau$} &
  $\rho$ &
  \multicolumn{1}{c|}{$\tau$} &
  $\rho$ &
  $\tau$ \\ \hline
L vs H &
  \multicolumn{1}{l|}{SL-TA vs SH-TA} &
  {\color[HTML]{111111} 0.441} &
  \multicolumn{1}{c|}{{\color[HTML]{111111} 0.393}} &
  {\color[HTML]{111111} 0.295} &
  \multicolumn{1}{c|}{0.739} &
  {\color[HTML]{111111} -0.914} &
  \multicolumn{1}{c|}{{\color[HTML]{111111} -0.328}} &
  {\color[HTML]{111111} -0.324} &
  \multicolumn{1}{c|}{{\color[HTML]{111111} 0.937}} &
  {\color[HTML]{111111} -0.133} &
  0.220 \\
 &
  \multicolumn{1}{l|}{ML-TA vs MH-TA} &
  {\color[HTML]{111111} 0.565} &
  \multicolumn{1}{c|}{{\color[HTML]{111111} 0.866}} &
  0.563 &
  \multicolumn{1}{c|}{{\color[HTML]{111111} 0.277}} &
  -0.295 &
  \multicolumn{1}{c|}{-0.409} &
  -0.456 &
  \multicolumn{1}{c|}{-0.455} &
  0.057 &
  0.228 \\ \hline
S vs M &
  \multicolumn{1}{l|}{SL-TA vs ML-TA} &
  {\color[HTML]{111111} 0.525} &
  \multicolumn{1}{c|}{{\color[HTML]{111111} 0.793}} &
  {\color[HTML]{111111} 0.989} &
  \multicolumn{1}{c|}{1.484} &
  0.385 &
  \multicolumn{1}{c|}{0.428} &
  {\color[HTML]{111111} 0.274} &
  \multicolumn{1}{c|}{{\color[HTML]{111111} 0.388}} &
  {\color[HTML]{111111} 0.528} &
  0.758 \\
 &
  \multicolumn{1}{l|}{SH-TA vs MH-TA} &
  {\color[HTML]{111111} 0.648} &
  \multicolumn{1}{c|}{{\color[HTML]{111111} 1.266}} &
  {\color[HTML]{111111} 1.257} &
  \multicolumn{1}{c|}{{\color[HTML]{111111} 1.022}} &
  {\color[HTML]{111111} 1.005} &
  \multicolumn{1}{c|}{{\color[HTML]{111111} 1.166}} &
  {\color[HTML]{111111} 0.142} &
  \multicolumn{1}{c|}{{\color[HTML]{111111} -0.145}} &
  {\color[HTML]{111111} 0.719} &
  {\color[HTML]{111111} 0.766} \\ \hline
S vs C &
  \multicolumn{1}{l|}{SL-TA vs Comb-TA} &
  {\color[HTML]{111111} 1.197} &
  \multicolumn{1}{c|}{{\color[HTML]{111111} 1.433}} &
  {\color[HTML]{111111} \textbf{3.216**}} &
  \multicolumn{1}{c|}{{\color[HTML]{111111} \textbf{3.54***}}} &
  {\color[HTML]{111111} \textbf{2.967**}} &
  \multicolumn{1}{c|}{{\color[HTML]{111111} \textbf{3.087**}}} &
  {\color[HTML]{111111} 1.318} &
  \multicolumn{1}{c|}{{\color[HTML]{111111} 1.602}} &
  {\color[HTML]{111111} 1.241} &
  {\color[HTML]{111111} 1.659} \\
 &
  \multicolumn{1}{l|}{SH-TA vs Comb-TA} &
  {\color[HTML]{111111} 0.755} &
  \multicolumn{1}{c|}{{\color[HTML]{111111} 1.040}} &
  \textbf{2.921**} &
  \multicolumn{1}{c|}{\textbf{2.808**}} &
  \textbf{3.881***} &
  \multicolumn{1}{c|}{\textbf{3.415***}} &
  1.643 &
  \multicolumn{1}{c|}{1.523} &
  1.374 &
  1.439 \\ \hline
M vs C &
  \multicolumn{1}{l|}{ML-TA vs Comb-TA} &
  {\color[HTML]{111111} 0.672} &
  \multicolumn{1}{c|}{{\color[HTML]{111111} 0.640}} &
  {\color[HTML]{111111} \textbf{2.226*}} &
  \multicolumn{1}{c|}{{\color[HTML]{111111} \textbf{2.063*}}} &
  \textbf{2.581**} &
  \multicolumn{1}{c|}{\textbf{2.659**}} &
  1.044 &
  \multicolumn{1}{c|}{1.213} &
  0.713 &
  0.901 \\
 &
  \multicolumn{1}{l|}{MH-TA vs Comb-TA} &
  {\color[HTML]{111111} 0.106} &
  \multicolumn{1}{c|}{{\color[HTML]{111111} -0.226}} &
  {\color[HTML]{111111} 1.663} &
  \multicolumn{1}{c|}{{\color[HTML]{111111} 1.786}} &
  \textbf{2.876**} &
  \multicolumn{1}{c|}{\textbf{2.249*}} &
  1.5 &
  \multicolumn{1}{c|}{1.669} &
  0.655 &
  0.672 \\ \hline
\end{tabular}%
}

\label{table4}
\end{table*}

\subsection{Comparative analysis of five TA conditions (\textbf{RQ}\ref{rq3})}

As this study differentiates the TA conditions, we observe how performance varies based on differences in attributes generated through the TA method under the five conditions: SL-TA, SH-TA, ML-TA, MH-TA, and Comb-TA. We then delve deeper into the effectiveness of combining the outcomes of humans' and LLMs' TA outcomes for generating checklists. The SL-TA score represents the correlation coefficients averaged across the four individual LLM conditions, and the SH-TA score indicates the coefficients averaged across the four different human expert conditions on the SummEval dataset.

\subsubsection{\textbf{Performance comparison across different TA conditions.}}

As shown in Figure \ref{fig4}, using the attributes generated by multiple humans or multiple LLMs results in better evaluation outcomes, including improved average performance, compared to using the ones produced by a single human or a single LLM, though the difference is not statistically significant (Table \ref{table4}). Moreover, Figure \ref{fig5} demonstrates that combining the TA outcomes from humans and LLMs for checklist construction (Comb-TA) outperforms all single-TA (SL-TA, SH-TA) and all multiple-TA (ML-TA, MH-TA) based checklists. While there is a minor exception in the Kendall's Tau score for Coherence, the gap is negligible and does not significantly affect the overall average performance. The improvement is significant in Fluency and Consistency (Table \ref{table4}), indicating that certain aspects of text quality benefit more from this combination than others.

\begin{figure*}[t]
    \centering
    \begin{subfigure}[b]{0.9\textwidth}
        \centering
        \includegraphics[width=0.9\textwidth]{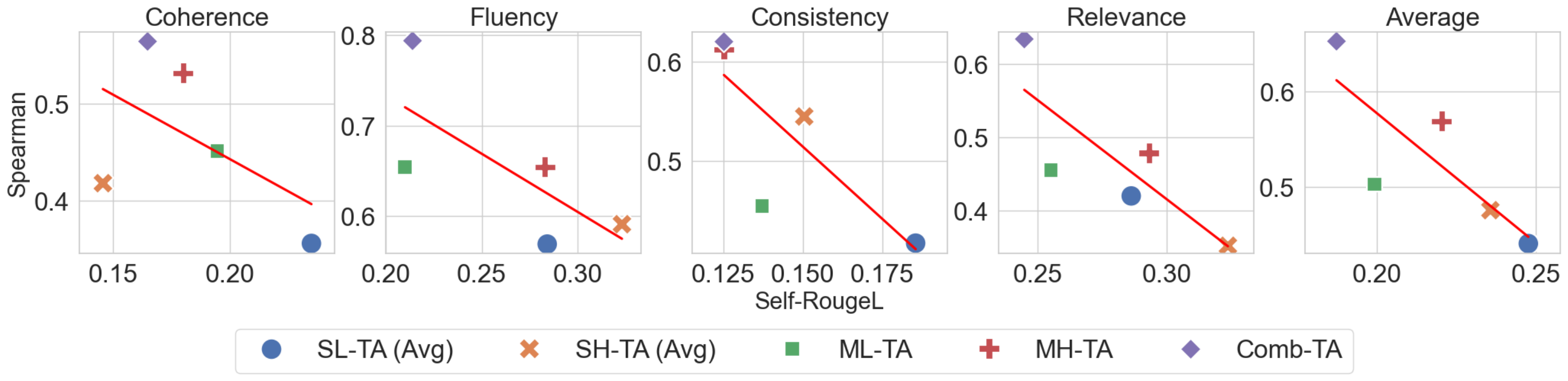}
        \caption{Comparison of evaluation performance. As Rouge-L (X-axis) decreases, Spearman's Rho ($\rho$) correlation (Y-axis) increases. The red line shows the upward trend across all dimensions. The decreasing Rouge-L signifies an increasing diversity of attributes within each condition.}
        \label{fig11}
    \end{subfigure}
    \begin{subfigure}[b]{0.9\textwidth}
        \centering
        \includegraphics[width=\textwidth]{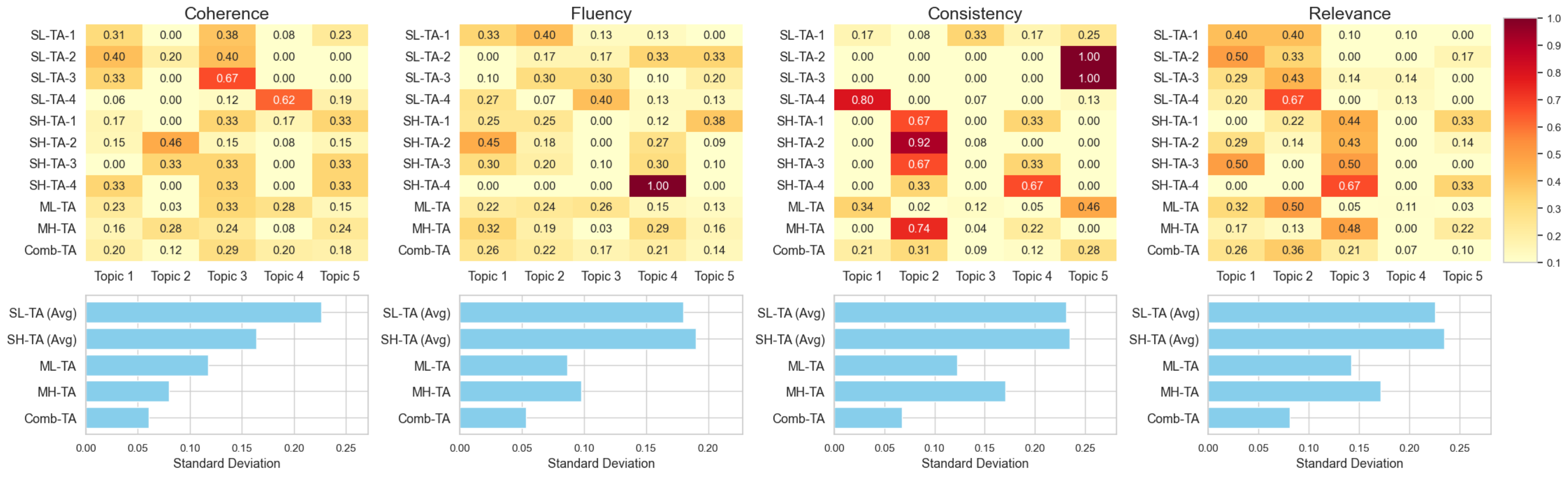}
        \caption{Topic Distribution and Standard Deviation Analysis using LDA: The upper figure shows that attributes from single-TA conditions tend to cluster around a few topics, whereas attributes from multiple-TA conditions are more evenly distributed across various topics. The lower figure indicates that the standard deviation of Comb-TA is the lowest, suggesting reduced bias.}
        \label{fig7}
    \end{subfigure}
    \caption{Analysis of Attributes and Evaluation Performance.}
    \label{fig6}
\end{figure*}

Figure \ref{fig11} shows the correlation coefficients between Spearman's Rho ($\rho$) values and self-ROUGE-L scores for attributes within the same TA condition. The results show that those checklists based on diverse attributes (i.e., with low similarity) enhance evaluation performance, supporting the idea that the multiple-TA condition helps create more fine-grained checklists and enhance the evaluation performance. Furthermore, combining human and LLM outcomes leads to even more fine-grained checklists, resulting in the highest performance and diversity.

To examine potential biases, particularly short-sighted or myopic cognitive biases, within the attributes of each TA condition, we conducted latent dirichlet allocation (LDA) \cite{lda}, as shown in Figure \ref{fig7}. This analysis generated five topics based on the attributes from the Comb-TA condition. The threshold of five is consistent with the maximum number of components extracted from each of the TA conditions. We then analyzed how the attributes from the TA conditions were concentrated in specific topics. As shown in Figure \ref{fig7}, the results show that attributes from the single-TA conditions, whether human or LLM, tend to cluster into a fewer topics, with some topics receiving no attributes at all, indicating the existence of myopic bias. In contrast, attributes from the multiple-TA and Comb-TA conditions are more evenly distributed across these five topics. Additionally, the Comb-TA condition has the lowest standard deviation in topic distribution, suggesting that combining the TA outcomes of multiple humans and LLMs results in a less biased checklist.

\begin{figure*}[t]
    \centering
    \includegraphics[width=\textwidth]{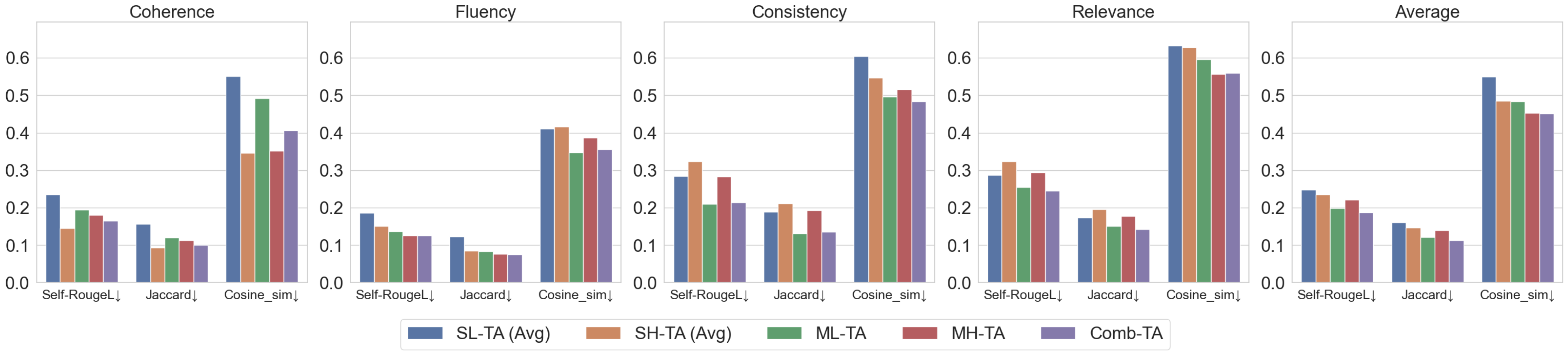}
    \caption{Similarity of Attributes in Lexical and Semantic Perspectives: The bars represent the measurement of text similarity. The similarity of SL-TA (Avg) and SH-TA (Avg) represents the average values across the four different checklists created by each LLM and each human expert, respectively. The lower the similarity, the more diverse the attributes are within a dimension.}
    \label{fig8}
\end{figure*}

\subsubsection{\textbf{Humans have strengths in internal quality}}\label{humans_strengths}

When comparing InteractEval's performance using checklists based on LLM-only TA (SL-TA and ML-TA) with human-only TA (SH-TA and MH-TA) in both single and multiple settings, human-only TA outperforms LLM-only TA in internal quality (Coherence and Fluency) of the generated texts (see Figure \ref{fig5}), suggesting that humans excel in identifying attributes related to structure, semantic connections, and readability, which are critical to internal text quality. Their flexible thinking skills \cite{blattberg} and innate understanding of universal grammar \cite{chomsky, universal} allow them to generate higher-quality attributes that account for the hierarchical structure of texts \cite{pantcheva}, a task that is more challenging for LLMs.

Qualitative analysis corroborates this finding. For ease of exposition, in this analysis, (A) refers to the attributes generated by humans and (B) refers to those generated by LLMs, during the TA process. For Coherence, humans created more quantitative and explicit attributes, making the qualitative analysis participants (E1-E4) easier to measure consistently. E3 commented, "(A) offers simple, straightforward evaluations with easily quantifiable items while (B) includes more qualitative, subjective items." Additionally, humans focused more on logical order and organization. E1 noted, "(A) emphasizes clear flows, meaningful connections, and natural sentence links for efficient information delivery while (B) focuses more on comprehension but is less efficient." For Fluency, humans provided attributes in more detail while LLMs took a broader approach. LLM-generated attributes sometimes confused qualitative study participants due to overlap with other dimensions. E1 remarked, "(A) offers detailed and specific requirements, like including at least two interrogatives while (B) provides more general guidelines." E2 noted, "(B) had many ambiguous items and concepts related to consistency." Additionally, the attribute similarities shown in Figure \ref{fig8} support these findings. Human-generated attributes for internal quality are less lexically and semantically similar compared to LLM-generated ones, suggesting that humans' TA produces more diverse attributes, enabling checklists to assess summaries across a wider range of features.

\subsubsection{\textbf{LLMs have strengths in external alignment.}}\label{llms_strengths}
In terms of external alignment (Consistency and Relevance), our framework using checklists based on LLM-only TA (SL-TA and ML-TA) shows superior performance or at least minimal differences, compared to those based on human-only TA (SH-TA and MH-TA), suggesting that LLMs excel in identifying attributes related to external alignment, where both the source text and the summary must be considered simultaneously. The task requires processing a large amount of information, where LLMs' vast context window and expansive memory can excel \cite{pantcheva, gpt4, wu2023reasoning}. LLMs are capable of providing a reliable set of attributes even when handling large data, adhering to their pre-trained policies \cite{blattberg, pantcheva}.

Qualitative analysis further supports these findings. For Consistency, LLMs explain the attributes in more detail and cover a broader scope than humans. E1 noted, "(B) describes each point with more detail," and E4 added, "(B) fully explains more details compared to (A)." Additionally, LLMs provide systematic guidelines for detecting hallucinations while human experts only provide attributes to check for false content. E2 commented, "(B) gives detailed guidelines for identifying hallucinations while (A) only shows how to detect them." For Relevance, humans tend to create confusing attributes that could be mapped to multiple dimensions. One expert mentioned being confused, thinking that certain attributes were Coherence: "In (A), some sentences could be interpreted as a coherence issue." E4 also noted, "Sometimes (A) felt like it belonged in another category." LLMs, on the other hand, clearly reflected the rubric's definition of Relevance and generated attributes covering a broader scope. Three out of four experts agreed on these points. E3 said, "(B) covers more areas and provides more accurate and detailed descriptions for each item, on the basis of the rubric."

As a result, LLM-generated attributes lead to better checklist construction and improved evaluation performance in the external alignment dimensions, as shown in Figure \ref{fig5}. Moreover, Figure \ref{fig8} shows that LLM-generated attributes for external alignment are less lexically and semantically similar compared to those created by humans, indicating that LLM-TA produces more diverse attributes, enabling checklists to assess summaries across a wider range of features.

\section{Discussion}
\label{discussion}
\subsection{Major Findings}
\begin{figure}[t]

        \centering
        \includegraphics[width=0.5\textwidth]{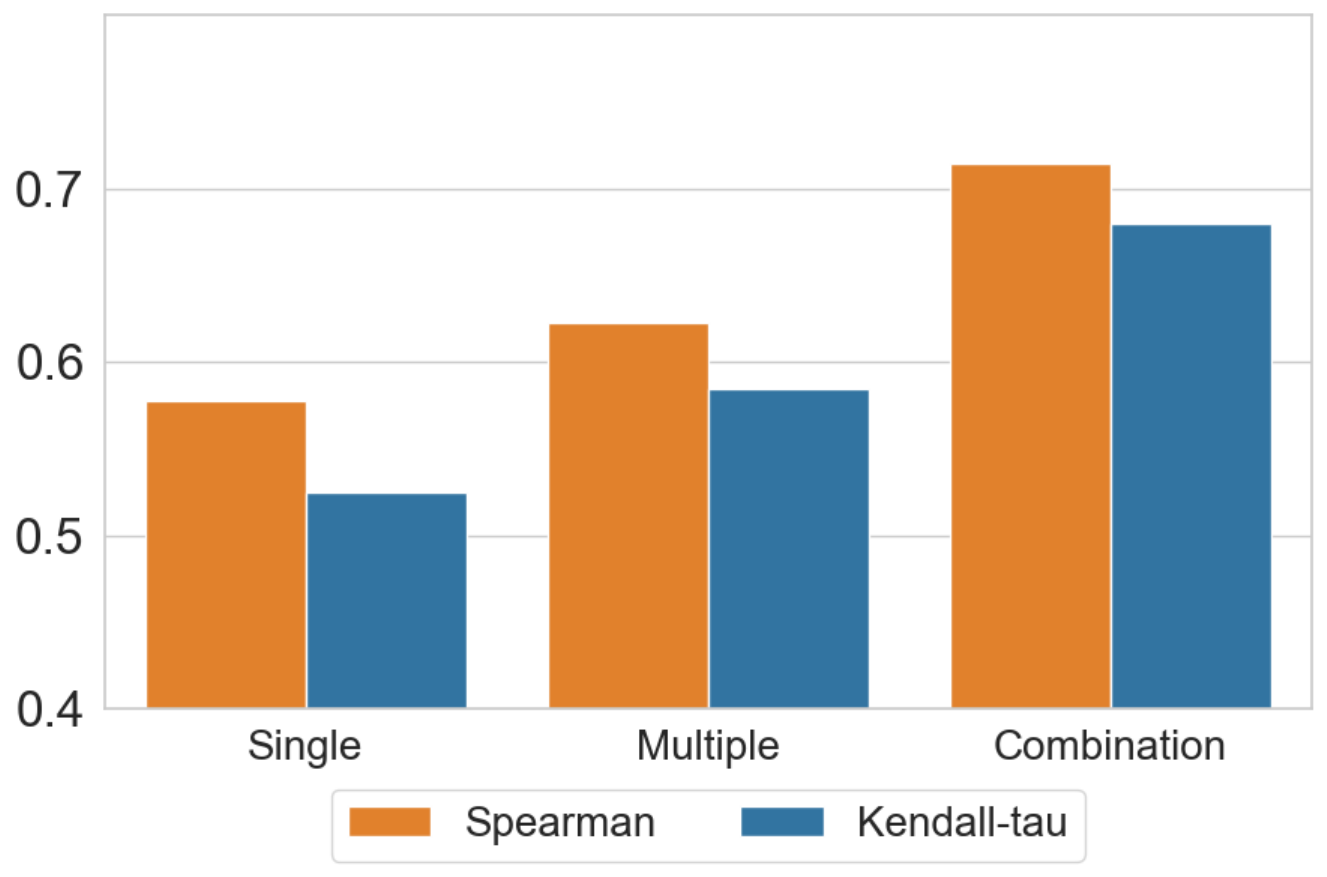}
 
    \caption{Comparison of evaluation performance averaged across four dimensions using two correlation metrics (Spearman's Rho ($\rho$) and Kendall's Tau ($\tau$)). Single refers to the average performance of InteractEval with checklists based on SL-TA and SH-TA, Multiple refers to the average performance with checklists based on ML-TA and MH-TA, and Combination represents the performance using checklists based on Comb-TA.}
    \label{fig9}
\end{figure}

In this study, we propose InteractEval, a text evaluation framework that uses checklists derived from attributes generated through Think Aloud (TA)  process based on human-machine collaboration. It assesses the framework's effectiveness compared to traditional and state-of-the-art baselines on a text summarization task, examines its transferability to a different task (essay scoring), and explores the potential synergistic effects of combining TA outcomes from humans and large language models (LLMs). Our key findings can be summarized as follows. First, InteractEval outperforms both traditional non-LLM and LLM-based baselines on evaluating summarized texts (Table \ref{table2}) and essays (Figure \ref{ellipse_performance}) in relation to correlating with the human-assigned ground-truth scores provided by the dataset. Particularly, analysis of attribute topics and similarities on the summary evaluation task shows that this improvement stems from reducing biases and enhancing fine-grainedness by incorporating both human and LLM perspectives, as well as rubrics and sample texts, supporting \textbf{RQ \ref{rq1}}. 

Secondly, the prompt engineering procedure employed for LLMs in this study, which is akin to the TA process employed for human experts, encourages LLMs to generate a wider range of text attributes, which form the basis of the checklists. This argument has been  confirmed by the strong performance of checklists based on the attributes from single-LLM TA, supporting \textbf{RQ \ref{rq2}}. In particular, considering that TA with each LLM demonstrates high performance in different dimensions, we conclude that combining the TA results from multiple LLMs complements each LLM's outcome and produces more reliable and refined checklists.

Finally, in response to \textbf{RQ \ref{rq3}}, combining the TA outcomes of humans and LLMs mitigates biased attributes and increases their fine-grainedness, leading to more refined checklists and improved evaluation performance. Specifically, humans and LLMs exhibit distinct strengths in text summary evaluation. Humans excel in assessing internal text quality, leveraging flexible thinking and an innate understanding of grammar. For example, attributes generated from human-only TA effectively capture the coherent and hierarchical structure of text \cite{pantcheva}, closely aligning with internal quality dimensions such as Coherence and Fluency. Consequently, more refined checklists can be designed through explicit and detailed attributes, resulting in more reliable evaluation performance. These findings highlight that human's empirical knowledge and practical skills surpass LLMs in capturing composition and organization features, such as semantic coherence and readability, consistent with a prior study \cite{dhillon}. In contrast, LLMs excel at detecting external alignment due to their ability to process large volumes of information consistently. Attributes generated from LLM-only TA effectively capture the semantic and structural connections between summary and source texts, closely related to external alignment dimensions such as Consistency and Relevance. These attributes improve checklist design by covering broader scopes and providing systematic guidelines for detecting hallucinations, underscoring LLMs' ability to efficiently manage large amounts of textual information and consistently compare two texts, which are in support of prior studies \cite{blattberg, wang2024}. Therefore, combining the outcomes of both humans' and LLMs' TA in checklist design leverages their complementary strengths, resulting in an optimal checklist and improved text evaluation performance for tasks such as text summarization (Figure \ref{fig9}), compared to using attributes from humans or LLMs alone. Additionally, the enhanced performance of InteractEval on the essay evaluation dataset underscores the transferability of its combinatory effect. This finding further highlights the advantages of human-AI collaboration, building on prior studies across various domains \cite{blattberg, wang2024, das, revilla, cohen}.

\subsection{Implications}
This study offers practical as well as research implications. As previously noted, text generation is rapidly expanding alongside advancements in LLMs. At the same time, the need for accurate methods to evaluate the quality of generated texts is gaining significant interest across various industry domains including education. InteractEval is a framework designed to answer the call. The framework can be utilized in various industry settings where the quality assessment of generated texts is critical. Similarly, for those research areas exploring the use of LLMs for text generation but lacking reliable metrics to evaluate the quality of the generated texts, our framework can serve as a useful testing tool. 
Moreover, our study empirically validates the impact of human-LLM collaboration during the think-aloud process, which constitutes the first stage of checklist design. By doing so, it suggests both the necessity and potential for future research to explore how collaboration between humans and LLMs in subsequent stages could further amplify this effect. Beyond the evaluation performance results, InteractEval demonstrates how humans and LLMs can work together to more reliably evaluate summaries and essays, anticipating that their synergistic potential will grow alongside advancements in LLM technology.

\section{Limitation and Future Work}
\label{limitations}
For this study, we have identified three main limitations. First, we tested our framework on a text summary evaluation task (SummEval \cite{summeval}) and an essay evaluation task (ELLIPSE \cite{ellipse}). However, our findings may not generalize to other types of text-related tasks, such as question-and-answering, image description, or text generation, as those tasks have different requirements for evaluation. Thus, further research is needed to explore the applicability of human-AI collaboration to text-related tasks beyond the evaluation of summaries and essays. Secondly, to ensure a fair comparison with the baselines that primarily use GPT models, we did not use smaller LLMs, such as Llama-3.1-8B-Instruct, as evaluators. Future studies could include the small-sized LLMs to assess the proposed methods' capabilities. Lastly, the two datasets used in this study are based in English, leaving room to examine the applicability of our approach to text evaluation tasks in other language settings.

\section{Conclusion}
\label{conclusion}
Our work distinguishes itself from other LLM-based text evaluation studies by paying attention to the potential synergetic effects that might be achieved by combining the strengths of humans and LLMs. To explore the understudied potential of human-AI collaboration in the context of text evaluation, we introduce InteractEval, a framework that integrates human expertise and LLMs using the concurrent think-aloud (TA) method to generate attributes for checklist-based text evaluation. Our findings demonstrate that combining the strengths of humans and LLMs significantly enhances text evaluation performance, highlighting the synergistic effect of their collaboration. Our study also emphasizes the effectiveness of TA in promoting divergent thinking in humans and divergent generation in LLMs, helping to identify a broader range of attributes, providing a clear explanation of the mechanism responsible for the superior performance outcomes observed. Furthermore, we empirically demonstrate that human experts and LLMs have their distinct but combinatory strengths. Humans excel in identifying attributes related to internal quality (Coherence and Fluency) whereas LLMs outperform in external alignment (Consistency and Relevance). Ultimately, as the title suggests, this study demonstrates that when humans and and LLMs collaborate, they can work better in text evaluation, opening up new horizons for effective human-machine collaboration.

\begin{acks}
This research was supported by Chunjae Education Inc. (Project Number G01250064).
\end{acks}

\bibliographystyle{ACM-Reference-Format}
\bibliography{ref}

\appendix
\section*{Appendix}

\section{Checklists of three rating dimensions in SummEval dataset (Table \ref{fluency_checklist}, \ref{consistency_checklist}, and \ref{relevance_checklist})}\label{appedix_checklist}

\section{Checklists of seven rating dimensions in ELLIPSE dataset (Table \ref{ellipse_overall_checklist}, \ref{ellipse_cohesion_checklist}, \ref{ellipse_conventions_checklist}, \ref{ellipse_grammar_checklist}, \ref{ellipse_phraseology_checklist}, \ref{ellipse_syntax_checklist}, and \ref{ellipse_vocabulary_checklist})}\label{appedix_checklist_ellipse}

\section{Information of human experts who participated in the TA process for ELLIPSE dataset (Table \ref{table_ellipse_ta_info})}\label{ellipse_ta_info}

\begin{figure}[h]
    \centering
    \includegraphics[width=\linewidth, height=0.6\linewidth]{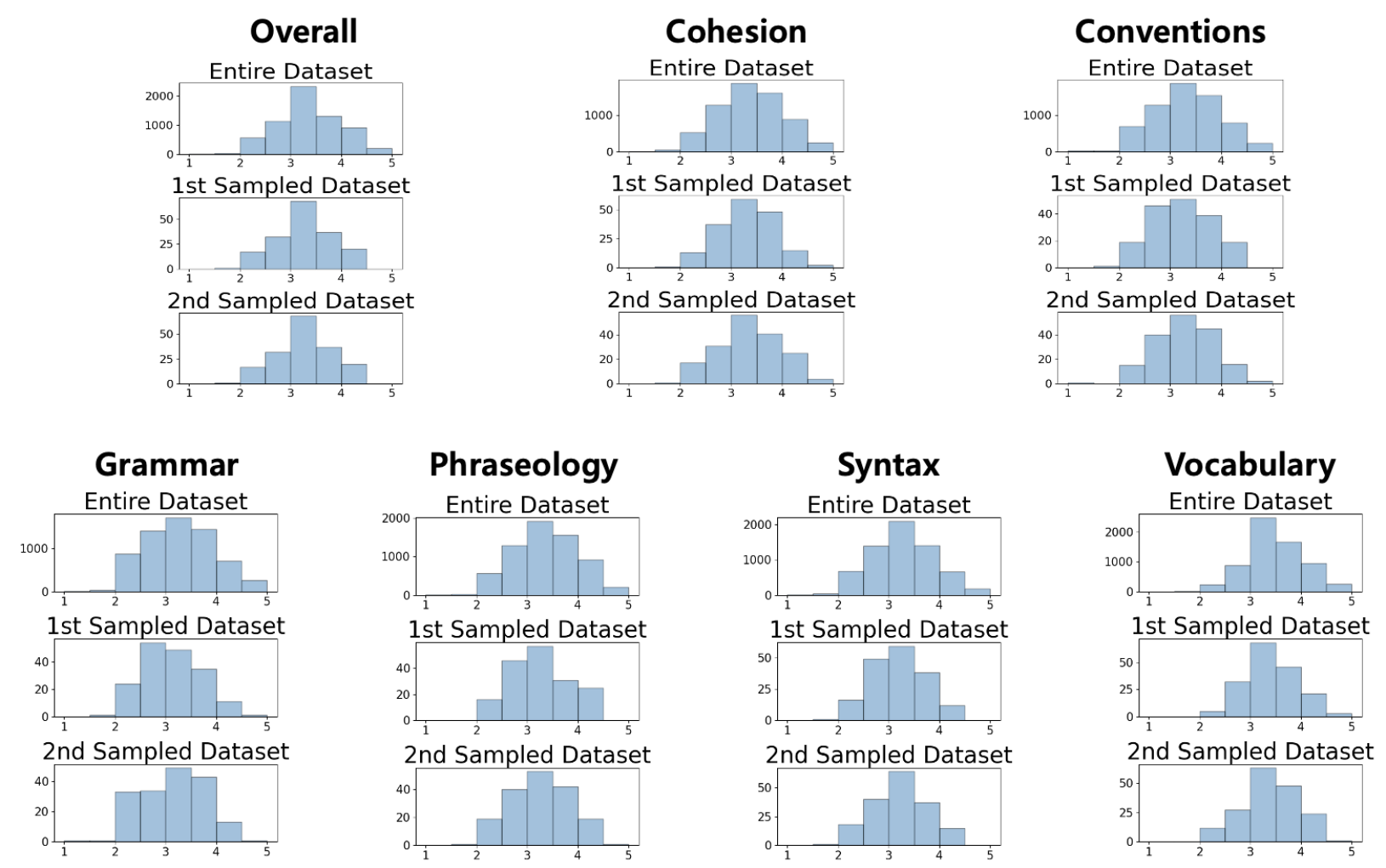}
    \caption{Comparison of distributions across seven dimensions. \textbf{Entire Dataset} refers to the distribution of the entire dataset with respect to each dimension, while \textbf{1st Sampled Dataset} and \textbf{2nd Sampled Dataset} represent the distributions of subsets obtained by sampling 10\% of the entire dataset.}
    \label{fig_ellipse_distribution}
\end{figure}

\section{Comparison of distributions in ELLIPSE dataset (Figure \ref{fig_ellipse_distribution}).}\label{distribution_of_ellipse}
Figure \ref{fig_ellipse_distribution} shows the score distributions for the entire ELLIPSE dataset, the first sampled dataset, and second sampled dataset across all dimensions. As shown, the sampled datasets preserve distributions nearly identical to those of the original dataset.

\section{Evaluation performance on ELLIPSE dataset (Table \ref{ellipse_performance_1} and \ref{ellipse_performance_2}).}\label{ellipse_performance_appendix}
Table \ref{ellipse_performance_1} and \ref{ellipse_performance_2} present the text evaluation performance on the ELLIPSE dataset. The table demonstrates that InteractEval's performance in both the first and second trials, using GPT-4, surpasses the baselines across all dimensions in terms of Spearman's Rho and Kendall's Tau correlations, except for the Spearman's Rho correlation in Cohesion.

\section{Open-ended questions for qualitative analysis}\label{open_ended_question}
Please answer the following questions and analyze the differences between Attributes (A) and Attribute (B).
\begin{itemize}
    \item What are the key points measured in “{\textit{Dimension}}” according to the given definition?
    \item Please compare and analyze (A) and (B) with respect to the evaluation of the “{\textit{Dimension}}.”
    \item Please compare and analyze (A) and (B) with respect to fine-grainedness.
    \item Please compare and analyze (A) and (B) regarding the diversity of their contents.
    \item Please compare and analyze (A) and (B), focusing on aspects not covered in the questions above.
    \item Please list the three most important differences between (A) and (B) overall. (You may repeat what has been already discussed.)
\end{itemize}

\newpage

\begin{table}[H]
\centering
\caption{Checklist of Fluency dataset}
\label{fluency_checklist}
\begin{tabularx}{\linewidth}{p{0.1\linewidth}X}
\hline
\multicolumn{1}{c}{\textbf{Dimension}} & \multicolumn{1}{c}{\textbf{Questions}} \\ \hline
Fluency & 
    - Does the summary follow proper capitalization rules? \newline
    - Is the formatting of the summary consistent and appropriate? \newline 
    - Does the summary avoid spelling errors or typos? \newline 
    - Are all sentences in the summary grammatically correct? \newline 
    - Does the summary use clear and concise language? \newline 
    - Does the summary effectively use transition words? \newline 
    - Does the summary effectively use both active and passive voice? \newline 
    - Does the summary flow smoothly from one idea to the next? \newline 
    - Is the tone of the summary appropriate for the content? \newline 
    - Does the style of the summary match the original document? \newline 
    - Is the summary concise without losing important information? \\ \hline
\end{tabularx}
\end{table}

\begin{table}[H]
\centering
\caption{Checklist of Consistency dataset}
\label{consistency_checklist}
\begin{tabularx}{\linewidth}{p{0.1\linewidth}X}
\hline
\multicolumn{1}{c}{\textbf{Dimension}} & \multicolumn{1}{c}{\textbf{Questions}} \\ \hline
Consistency & 
    - Does the summary correctly represent the key facts from the source document? \newline 
    - Is the summary free from misrepresentation of the facts in the source document? \newline 
    - Does the summary maintain the same factual context as the source document? \newline 
    - Do the statements in the summary logically follow from the information in the source document? \newline 
    - Is the summary free from bias or subjective interpretations of the source document? \newline 
    - Does the summary refrain from introducing new facts that were not mentioned in the source document? \newline 
    - Are all the important facts in the summary traceable to the source document?\\ \hline
\end{tabularx}
\end{table}

\begin{table}[H]
\centering
\caption{Checklist of Relevance dataset}
\label{relevance_checklist}
\begin{tabularx}{\linewidth}{p{0.1\linewidth}X}
\hline
\multicolumn{1}{c}{\textbf{Dimension}} & \multicolumn{1}{c}{\textbf{Questions}} \\ \hline
Relevance & 
    - Does the summary include all the crucial details from the source text? \newline 
    - Does the summary exclude insignificant details from the source text? \newline 
    - Is the summary logically organized? \newline 
    - Does the summary have a clear and understandable structure? \newline 
    - Are the events from the source text accurately represented in the summary? \newline 
    - Does the summary maintain the context of the original document? \newline 
    - Does the summary avoid including personal interpretations or opinions? \newline 
    - Is the summary relevant to the original article's purpose? \newline 
    - Does the summary effectively convey the original article's main points to a reader who has not read the original article? \newline 
    - Is the length of the summary appropriate in relation to the original article? \newline 
    - Does the summary avoid unnecessary wordiness or repetition?\\ \hline
\end{tabularx}
\end{table}

\begin{table}[H]
\centering
\caption{Checklist of Overall dimension}
\label{ellipse_overall_checklist}
\begin{tabularx}{\linewidth}{p{0.1\linewidth}X}
\hline
\multicolumn{1}{c}{\textbf{Dimension}} & \multicolumn{1}{c}{\textbf{Questions}} \\ \hline
Overall & 
    - Does the essay incorporate a variety of vocabulary that is relevant to the topic? \newline
    - Are the words and phrases used in the essay contextually appropriate? \newline 
    - Does the essay exhibit a high level of grammatical accuracy? \newline 
    - Are complex grammatical structures used correctly in the essay? \newline 
    - Does the essay have a clear and logical structure? \newline 
    - Are the ideas in the essay developed in a coherent and systematic manner? \newline 
    - Does the essay maintain a consistent focus on the central theme? \newline 
    - Does the essay demonstrate a high level of control over sentence structure? \newline 
    - Are the sentences in the essay well-constructed and varied in length and structure? \newline 
    - Does the essay maintain a logical and clear organization throughout? \newline 
    - Does the essay maintain effective communication despite any language inaccuracies? \\ \hline
\end{tabularx}
\end{table}


\begin{table}[H]
\centering
\caption{Checklist of Cohesion dimension}
\label{ellipse_cohesion_checklist}
\begin{tabularx}{\linewidth}{p{0.1\linewidth}X}
\hline
\multicolumn{1}{c}{\textbf{Dimension}} & \multicolumn{1}{c}{\textbf{Questions}} \\ \hline
Cohesion & 
    - Does the essay incorporate transitional words to connect ideas? \newline 
    - Are phrases used effectively to link concepts in the essay? \newline 
    - Does the essay use reference words to connect ideas? \newline 
    - Are the paragraphs in the essay interconnected through overlapping ideas? \newline 
    - Does the essay maintain a consistent flow of ideas? \newline 
    - Is the organization of the essay logical and easy to understand? \newline 
    - Does the essay avoid repetitive use of the same cohesive devices? \newline 
    - Are the cohesive devices used in the essay accurate and appropriate for the context? \newline 
    - Are the ideas in the essay presented in a clear and concise manner? \\ \hline
\end{tabularx}
\end{table}

\begin{table}[H]
\centering
\caption{Checklist of Conventions dimension}
\label{ellipse_conventions_checklist}
\begin{tabularx}{\linewidth}{p{0.1\linewidth}X}
\hline
\multicolumn{1}{c}{\textbf{Dimension}} & \multicolumn{1}{c}{\textbf{Questions}} \\ \hline
Conventions & 
    - Does the essay incorporate transitional words to link sentences and paragraphs? \newline 
    - Are reference words used effectively to avoid repetition? \newline 
    - Does the essay use other linguistic features to connect ideas? \newline 
    - Does the essay have a clear introduction, body, and conclusion? \newline 
    - Are the ideas presented in a logical sequence? \newline 
    - Does the essay maintain a consistent control of ideas from beginning to end? \newline 
    - Are the ideas in the essay expressed in a clear and understandable manner? \newline 
    - Does the essay use clear language and avoid jargon? \newline 
    - Are relevant literary devices used to enhance the presentation of ideas? \newline 
    - Are all words in the essay spelled correctly? \newline 
    - Does the essay use punctuation marks correctly to enhance readability? \newline 
    - Are all sentences properly capitalized? \\ \hline
\end{tabularx}
\end{table}

\begin{table}[H]
\centering
\caption{Checklist of Grammar dimension}
\label{ellipse_grammar_checklist}
\begin{tabularx}{\linewidth}{p{0.1\linewidth}X}
\hline
\multicolumn{1}{c}{\textbf{Dimension}} & \multicolumn{1}{c}{\textbf{Questions}} \\ \hline
Grammar & 
    - Does the essay consistently use correct spelling? \newline 
    - Is capitalization correctly and consistently applied throughout the essay? \newline 
    - Does the essay demonstrate correct and consistent use of punctuation? \newline 
    - Does the essay demonstrate a strong command of grammar with minimal errors? \newline 
    - Does the essay effectively use a variety of sentence structures? \newline 
    - Does the essay consistently use appropriate vocabulary? \newline 
    - Are transitions effectively used throughout the essay? \newline 
    - Does the essay demonstrate coherence in its argument or narrative? \newline 
    - Does the essay effectively use rhetorical devices? \\ \hline
\end{tabularx}
\end{table}

\begin{table}[H]
\centering
\caption{Checklist of Phraseology dimension}
\label{ellipse_phraseology_checklist}
\begin{tabularx}{\linewidth}{p{0.1\linewidth}X}
\hline
\multicolumn{1}{c}{\textbf{Dimension}} & \multicolumn{1}{c}{\textbf{Questions}} \\ \hline
Phraseology & 
    - Does the essay refrain from overusing memorized chunks of language? \newline 
    - Is there a low occurrence of misuse of simple phrasal patterns in the essay? \newline 
    - Does the essay incorporate a diverse range of phrases? \newline 
    - Are idioms, collocations, and lexical bundles used effectively in the essay? \newline 
    - Does the essay avoid the overuse of a single type of phrase, such as idioms, collocations, or lexical bundles? \newline 
    - Are the phrases used in the essay suitable for the topic? \newline 
    - Do the phrases contribute effectively to the overall argument or point of the essay? \newline 
    - Does the essay demonstrate flexibility in the use of phrases? \newline 
    - Are the phrases used in the essay able to convey precise meanings? \newline 
    - Does the essay use phrases that convey subtle meanings effectively? \newline 
    - Are inaccuracies in phrase usage infrequent and negligible in the essay? \\ \hline
\end{tabularx}
\end{table}

\begin{table}[H]
\centering
\caption{Checklist of Syntax dimension}
\label{ellipse_syntax_checklist}
\begin{tabularx}{\linewidth}{p{0.1\linewidth}X}
\hline
\multicolumn{1}{c}{\textbf{Dimension}} & \multicolumn{1}{c}{\textbf{Questions}} \\ \hline
Syntax & 
    - Does the essay maintain correct word order in sentences? \newline 
    - Does the essay avoid sentence fragments and run-on sentences? \newline 
    - Does the essay include a mix of simple, compound, and complex sentences? \newline 
    - Are the sentence structures used effectively to convey the intended meaning? \newline 
    - Does the essay avoid overuse of any particular sentence structure? \newline 
    - Are any sentence formation errors present minor and infrequent? \newline 
    - Does the essay use a variety of syntactic structures? \newline 
    - Are the syntactic structures used appropriate for the context and content of the essay? \\ \hline
\end{tabularx}
\end{table}

\begin{table}[H]
\centering
\caption{Checklist of Vocabulary dimension}
\label{ellipse_vocabulary_checklist}
\begin{tabularx}{\linewidth}{p{0.1\linewidth}X}
\hline
\multicolumn{1}{c}{\textbf{Dimension}} & \multicolumn{1}{c}{\textbf{Questions}} \\ \hline
Vocabulary & 
    - Does the essay use a variety of different words and phrases? \newline 
    - Does the essay use advanced vocabulary where appropriate? \newline 
    - Does the vocabulary used in the essay accurately convey the intended meanings? \newline 
    - Does the essay use formal language where required? \newline 
    - Does the vocabulary used match the context and purpose of the essay? \newline 
    - Are the topic-related terms used accurately within the context of the essay? \newline 
    - Does the essay use correct grammatical forms of words? \newline 
    - Does the essay use less common words and phrases accurately? \newline 
    - Does the use of less common lexical items demonstrate a high level of language proficiency? \\ \hline
\end{tabularx}
\end{table}

\begin{table*}[t]
\centering
\caption{Information of human experts who participated in the TA process.}

\resizebox{0.8\textwidth}{!}{%
\begin{tabular}{llll}
\hline
Expert & Occupation & Related Experience & Language Use (English) \\ \hline
1 & Ph.D Student & Researching on applied linguistics & Fluent (non-native) \\
2 & University Lecturer & Teaching English writing in a university & Fluent (native) \\
3 & University Lecturer & Teaching English writing in a university & Fluent (native) \\
4 & University Lecturer & Teaching English writing in a university & Fluent (native) \\ \hline
\end{tabular}%
}
\label{table_ellipse_ta_info}
\end{table*}

\begin{table*}[t]
\centering
\caption{The first trial of sample-level evaluation performance of different dimensions on ELLIPSE benchmark dataset measured by Spearman's Rho ($\rho$), Kendall's Tau ($\tau$), and Mean Absolute Error (\textit{m}). The best results are highlighted in \textbf{bold}, and the second-best results are \underline{underlined}.}
\label{ellipse_performance_1}
\resizebox{\textwidth}{!}{%
\begin{tabular}{lllllllll}
\hline
\multicolumn{1}{c|}{{\textbf{Model}}} &
  \multicolumn{1}{c|}{{\textbf{Overall}}} &
  \multicolumn{1}{c|}{{\textbf{Cohesion}}} &
  \multicolumn{1}{c|}{{\textbf{Conventions}}} &
  \multicolumn{1}{c|}{{\textbf{Grammar}}} &
  \multicolumn{1}{c|}{{\textbf{Phraseology}}} &
  \multicolumn{1}{c|}{{\textbf{Syntax}}} &
  \multicolumn{1}{c|}{{\textbf{Vocabulary}}} &
  \multicolumn{1}{c}{{\textbf{Average}}} \\ 
  \cline{2-9} 
\multicolumn{1}{l|}{{}} &
  \multicolumn{1}{c|}{{$\rho \uparrow$ / $\tau \uparrow$ (\textit{m}$\downarrow$)}} &
  \multicolumn{1}{c|}{{$\rho \uparrow$ / $\tau \uparrow$ (\textit{m}$\downarrow$)}} &
  \multicolumn{1}{c|}{{$\rho \uparrow$ / $\tau \uparrow$ (\textit{m}$\downarrow$)}} &
  \multicolumn{1}{c|}{{$\rho \uparrow$ / $\tau \uparrow$ (\textit{m}$\downarrow$)}} &
  \multicolumn{1}{c|}{{$\rho \uparrow$ / $\tau \uparrow$ (\textit{m}$\downarrow$)}} &
  \multicolumn{1}{c|}{{$\rho \uparrow$ / $\tau \uparrow$ (\textit{m}$\downarrow$)}} &
  \multicolumn{1}{c|}{{$\rho \uparrow$ / $\tau \uparrow$ (\textit{m}$\downarrow$)}} &
  \multicolumn{1}{c}{{$\rho \uparrow$ / $\tau \uparrow$ (\textit{m}$\downarrow$)}} \\ \hline
{ \textbf{Non-LLM based}} &
  {} &
  {} &
  {} &
  {} &
  {} &
  {} &
  {} &
  {} \\ \hline
\multicolumn{1}{l|}{{ROUGE-L}} &
  \multicolumn{1}{l|}{{0.084/0.062 (2.866)}} &
  \multicolumn{1}{l|}{{0.287/0.211 (2.892)}} &
  \multicolumn{1}{l|}{{0.166/0.118 (2.816)}} &
  \multicolumn{1}{l|}{{0.126/0.011 (2.714)}} &
  \multicolumn{1}{l|}{{0.222/0.013 (2.856)}} &
  \multicolumn{1}{l|}{{0.206/0.152 (2.778)}} &
  \multicolumn{1}{l|}{{0.142/0.107 (3.000)}} &
  {0.176/0.096 (2.846)} \\
\multicolumn{1}{l|}{{BLEU}} &
  \multicolumn{1}{l|}{{0.306/0.229 (2.941)}} &
  \multicolumn{1}{l|}{{0.303/0.225 (2.970)}} &
  \multicolumn{1}{l|}{{0.250/0.186 (2.875)}} &
  \multicolumn{1}{l|}{{0.271/0.200 (2.773)}} &
  \multicolumn{1}{l|}{{0.235/0.171 (2.936)}} &
  \multicolumn{1}{l|}{{0.243/0.179 (2.848)}} &
  \multicolumn{1}{l|}{{0.291/0.221 (3.069)}} &
  {0.271/0.202 (2.916)} \\
\multicolumn{1}{l|}{{METEOR}} &
  \multicolumn{1}{l|}{{0.331/0.254 (2.037)}} &
  \multicolumn{1}{l|}{{0.339/0.255 (2.070)}} &
  \multicolumn{1}{l|}{{0.246/0.185 (1.950)}} &
  \multicolumn{1}{l|}{{0.216/0.162 (1.849)}} &
  \multicolumn{1}{l|}{{0.336/0.259 (2.028)}} &
  \multicolumn{1}{l|}{{0.234/0.173 (1.907)}} &
  \multicolumn{1}{l|}{{0.345/0.264 (2.162)}} &
  {0.292/0.222 (2.000)} \\
\multicolumn{1}{l|}{{BERTScore}} &
  \multicolumn{1}{l|}{{0.508/0.390 (1.078)}} &
  \multicolumn{1}{l|}{{0.506/0.384 (1.060)}} &
  \multicolumn{1}{l|}{{0.435/0.327 (1.132)}} &
  \multicolumn{1}{l|}{{0.464/0.346 (1.241)}} &
  \multicolumn{1}{l|}{{0.443/0.333 (1.085)}} &
  \multicolumn{1}{l|}{{0.482/0.365 (1.162)}} &
  \multicolumn{1}{l|}{{0.436/0.338 (0.953)}} &
  {0.468/0.355 (1.102)} \\
\multicolumn{1}{l|}{{MOVERScore}} &
  \multicolumn{1}{l|}{{0.335/0.287 (1.854)}} &
  \multicolumn{1}{l|}{{0.368/0.294 (1.648)}} &
  \multicolumn{1}{l|}{{0.351/0.254 (1.514)}} &
  \multicolumn{1}{l|}{{0.387/0.329 (1.631)}} &
  \multicolumn{1}{l|}{{0.325/0.209 (1.593)}} &
  \multicolumn{1}{l|}{{0.377/0.313 (1.498)}} &
  \multicolumn{1}{l|}{{0.401/0.292 (\underline{0.952})}} &
  {0.363/0.283 (1.527)} \\
\multicolumn{1}{l|}{{BARTScore}} &
  \multicolumn{1}{l|}{{-0.010/-0.008 (\underline{0.850})}} &
  \multicolumn{1}{l|}{{-0.009/-0.003 (\underline{0.885})}} &
  \multicolumn{1}{l|}{{-0.012/-0.011 (0.858)}} &
  \multicolumn{1}{l|}{{0.016/0.013 (0.885)}} &
  \multicolumn{1}{l|}{{-0.002/-0.003 (0.837)}} &
  \multicolumn{1}{l|}{{0.033/0.024 (0.933)}} &
  \multicolumn{1}{l|}{{0.126/0.094 (1.040)}} &
  {0.020/0.015 (0.898)} \\
\multicolumn{1}{l|}{{UniEval}} &
  \multicolumn{1}{l|}{{0.613/0.476 (1.780)}} &
  \multicolumn{1}{l|}{{0.551/0.422 (1.118)}} &
  \multicolumn{1}{l|}{{0.521/0.438 (1.676)}} &
  \multicolumn{1}{l|}{{0.508/\underline{0.546} (1.746)}} &
  \multicolumn{1}{l|}{{0.587/0.459 (1.435)}} &
  \multicolumn{1}{l|}{{0.505/\underline{0.470} (1.735)}} &
  \multicolumn{1}{l|}{{0.625/0.533 (1.490)}} &
  {0.559/0.478 (1.569)} \\ \hline
{\textbf{LLM based (GPT 3.5-Turbo)}} &
  {} &
  {} &
  {} &
  {} &
  {} &
  {} &
  {} &
  {} \\ \hline
\multicolumn{1}{l|}{{G-Eval}} &
  \multicolumn{1}{l|}{{0.529/0.421 (1.005)}} &
  \multicolumn{1}{l|}{{0.470/0.364 (0.951)}} &
  \multicolumn{1}{l|}{{0.425/0.328 (\underline{0.810})}} &
  \multicolumn{1}{l|}{{0.346/0.265 (\underline{0.811})}} &
  \multicolumn{1}{l|}{{0.534/0.419 (\underline{0.750})}} &
  \multicolumn{1}{l|}{{0.494/0.388 (\underline{0.649})}} &
  \multicolumn{1}{l|}{{0.537/0.425 (0.998)}} &
  {0.476/0.373 (\textbf{0.853})} \\
\multicolumn{1}{l|}{{CheckEval}} &
  \multicolumn{1}{l|}{{0.424/0.380 (1.923)}} &
  \multicolumn{1}{l|}{{0.313/0.288 (1.879)}} &
  \multicolumn{1}{l|}{{0.128/0.118 (2.002)}} &
  \multicolumn{1}{l|}{{-0.166/-0.152 (1.663)}} &
  \multicolumn{1}{l|}{{0.399/0.352 (1.809)}} &
  \multicolumn{1}{l|}{{0.509/0.440 (1.562)}} &
  \multicolumn{1}{l|}{{0.593/0.522 (1.622)}} &
  {0.314/0.278 (1.780)} \\
\multicolumn{1}{l|}{{InteractEval}} &
  \multicolumn{1}{l|}{{0.432/0.387 (1.863)}} &
  \multicolumn{1}{l|}{{0.416/0.373 (1.788)}} &
  \multicolumn{1}{l|}{{0.376/0.336 (1.829)}} &
  \multicolumn{1}{l|}{{0.509/0.449 (1.616)}} &
  \multicolumn{1}{l|}{{0.542/0.452 (1.386)}} &
  \multicolumn{1}{l|}{{0.523/0.452 (1.018)}} &
  \multicolumn{1}{l|}{{0.577/0.513 (1.527)}} &
  {0.482/0.423 (1.575)} \\ \hline
{\textbf{LLM based (GPT-4)}} &
  {} &
  {} &
  {} &
  {} &
  {} &
  {} &
  {} &
  {} \\ \hline
\multicolumn{1}{l|}{{G-Eval}} &
  \multicolumn{1}{l|}{{0.645/0.522 (0.912)}} &
  \multicolumn{1}{l|}{{\textbf{0.570}/0.461 (\textbf{0.866})}} &
  \multicolumn{1}{l|}{{0.541/\underline{0.451} (0.853)}} &
  \multicolumn{1}{l|}{{\underline{0.618}/0.491 (0.971)}} &
  \multicolumn{1}{l|}{{0.549/0.430 (1.170)}} &
  \multicolumn{1}{l|}{{0.542/0.433 (1.038)}} &
  \multicolumn{1}{l|}{{0.608/0.481 (1.028)}} &
  {\underline{0.582}/0.467 (0.977)} \\
\multicolumn{1}{l|}{{CheckEval}} &
  \multicolumn{1}{l|}{{\underline{0.657}/\underline{0.546} (0.999)}} &
  \multicolumn{1}{l|}{{0.561/\underline{0.462} (1.256)}} &
  \multicolumn{1}{l|}{{\underline{0.547}/0.445 (0.922)}} &
  \multicolumn{1}{l|}{{0.481/0.424 (\textbf{0.562})}} &
  \multicolumn{1}{l|}{{\underline{0.598}/\underline{0.506} (1.319)}} &
  \multicolumn{1}{l|}{{\underline{0.555}/0.463 (1.243)}} &
  \multicolumn{1}{l|}{{\underline{0.642}/\underline{0.539} (1.261)}} &
  {0.577/\underline{0.484} (1.080)} \\
\multicolumn{1}{l|}{{InteractEval}} &
  \multicolumn{1}{l|}{{\textbf{0.679}/\textbf{0.580} (\textbf{0.771})}} &
  \multicolumn{1}{l|}{{\underline{0.564}/\textbf{0.486} (1.404)}} &
  \multicolumn{1}{l|}{{\textbf{0.557}/\textbf{0.453} (\textbf{0.757})}} &
  \multicolumn{1}{l|}{{\textbf{0.654}/\textbf{0.550} (0.906)}} &
  \multicolumn{1}{l|}{{\textbf{0.669}/\textbf{0.575} (\textbf{0.736})}} &
  \multicolumn{1}{l|}{{\textbf{0.599}/\textbf{0.509} (\textbf{0.634})}} &
  \multicolumn{1}{l|}{{\textbf{0.660}/\textbf{0.568} (\textbf{0.910})}} &
  {\textbf{0.626}/\textbf{0.532} (\underline{0.874})} \\ \hline
\end{tabular}%
}
\end{table*}

\begin{table*}[t]
\centering
\caption{The second trial of sample-level evaluation performance of different dimensions on ELLIPSE benchmark dataset measured by Spearman's Rho ($\rho$), Kendall's Tau ($\tau$), and Mean Absolute Error (\textit{m}). The best results are highlighted in \textbf{bold}, and the second-best results are \underline{underlined}.}
\label{ellipse_performance_2}
\resizebox{\textwidth}{!}{%
\begin{tabular}{lllllllll}
\hline
\multicolumn{1}{c|}{{\textbf{Model}}} &
  \multicolumn{1}{c|}{{\textbf{Overall}}} &
  \multicolumn{1}{c|}{{\textbf{Cohesion}}} &
  \multicolumn{1}{c|}{{\textbf{Conventions}}} &
  \multicolumn{1}{c|}{{\textbf{Grammar}}} &
  \multicolumn{1}{c|}{{\textbf{Phraseology}}} &
  \multicolumn{1}{c|}{{\textbf{Syntax}}} &
  \multicolumn{1}{c|}{{\textbf{Vocabulary}}} &
  \multicolumn{1}{c}{{\textbf{Average}}} \\ \cline{2-9} 
\multicolumn{1}{l|}{{}} &
  \multicolumn{1}{c|}{{$\rho \uparrow$ / $\tau \uparrow$ (\textit{m}$\downarrow$)}} &
  \multicolumn{1}{c|}{{$\rho \uparrow$ / $\tau \uparrow$ (\textit{m}$\downarrow$)}} &
  \multicolumn{1}{c|}{{$\rho \uparrow$ / $\tau \uparrow$ (\textit{m}$\downarrow$)}} &
  \multicolumn{1}{c|}{{$\rho \uparrow$ / $\tau \uparrow$ (\textit{m}$\downarrow$)}} &
  \multicolumn{1}{c|}{{$\rho \uparrow$ / $\tau \uparrow$ (\textit{m}$\downarrow$)}} &
  \multicolumn{1}{c|}{{$\rho \uparrow$ / $\tau \uparrow$ (\textit{m}$\downarrow$)}} &
  \multicolumn{1}{c|}{{$\rho \uparrow$ / $\tau \uparrow$ (\textit{m}$\downarrow$)}} &
  \multicolumn{1}{c}{{$\rho \uparrow$ / $\tau \uparrow$ (\textit{m}$\downarrow$)}} \\ \hline
{\textbf{Non-LLM based}} &
  {} &
  {} &
  {} &
  {} &
  {} &
  {} &
  {} &
  {} \\ \hline
\multicolumn{1}{l|}{{ROUGE-L}} &
  \multicolumn{1}{l|}{{0.144/0.107 (2.866)}} &
  \multicolumn{1}{l|}{{0.248/0.184 (2.939)}} &
  \multicolumn{1}{l|}{{0.146/0.108 (2.870)}} &
  \multicolumn{1}{l|}{{0.021/0.016 (2.742)}} &
  \multicolumn{1}{l|}{{0.030/0.024 (2.855)}} &
  \multicolumn{1}{l|}{{0.209/0.153 (2.806)}} &
  \multicolumn{1}{l|}{{0.200/0.149 (2.979)}} &
  {0.143/0.106 (2.865)} \\
\multicolumn{1}{l|}{{BLEU}} &
  \multicolumn{1}{l|}{{0.274/0.204 (2.944)}} &
  \multicolumn{1}{l|}{{0.227/0.166 (3.018)}} &
  \multicolumn{1}{l|}{{0.026/0.022 (2.932)}} &
  \multicolumn{1}{l|}{{0.184/0.134 (2.803)}} &
  \multicolumn{1}{l|}{{0.187/0.142 (2.931)}} &
  \multicolumn{1}{l|}{{0.161/0.120 (2.875)}} &
  \multicolumn{1}{l|}{{0.136/0.102 (3.049)}} &
  {0.171/0.127 (2.936)} \\
\multicolumn{1}{l|}{{METEOR}} &
  \multicolumn{1}{l|}{{0.281/0.210 (2.046)}} &
  \multicolumn{1}{l|}{{0.349/0.261 (2.128)}} &
  \multicolumn{1}{l|}{{0.147/0.107 (2.018)}} &
  \multicolumn{1}{l|}{{0.117/0.087 (1.886)}} &
  \multicolumn{1}{l|}{{0.324/0.243 (2.029)}} &
  \multicolumn{1}{l|}{{0.256/0.193 (1.945)}} &
  \multicolumn{1}{l|}{{0.298/0.231 (2.146)}} &
  {0.253/0.190 (2.028)} \\
\multicolumn{1}{l|}{{BERTScore}} &
  \multicolumn{1}{l|}{{0.581/0.452 (1.076)}} &
  \multicolumn{1}{l|}{{0.476/0.356 (1.018)}} &
  \multicolumn{1}{l|}{{0.416/0.319 (1.083)}} &
  \multicolumn{1}{l|}{{0.455/0.337 (1.211)}} &
  \multicolumn{1}{l|}{{0.510/0.393 (1.092)}} &
  \multicolumn{1}{l|}{{0.541/0.462 (1.132)}} &
  \multicolumn{1}{l|}{{0.554/0.433 (0.964)}} &
  {0.505/0.393 (1.082)} \\
\multicolumn{1}{l|}{{MOVERScore}} &
  \multicolumn{1}{l|}{{0.365/0.312 (1.732)}} &
  \multicolumn{1}{l|}{{0.387/0.318 (1.547)}} &
  \multicolumn{1}{l|}{{0.372/0.274 (1.612)}} &
  \multicolumn{1}{l|}{{0.398/0.342 (1.472)}} &
  \multicolumn{1}{l|}{{0.342/0.223 (1.485)}} &
  \multicolumn{1}{l|}{{0.403/0.322 (1.432)}} &
  \multicolumn{1}{l|}{{0.435/0.364 (0.954)}} &
  {0.386/0.308 (1.462)} \\
\multicolumn{1}{l|}{{BARTScore}} &
  \multicolumn{1}{l|}{{0.036/0.028 (\underline{0.847})}} &
  \multicolumn{1}{l|}{{0.097/0.073 (\underline{0.936})}} &
  \multicolumn{1}{l|}{{0.100/0.074 (\textbf{0.860})}} &
  \multicolumn{1}{l|}{{-0.012/-0.012 (0.954)}} &
  \multicolumn{1}{l|}{{-0.072/-0.053 (0.877)}} &
  \multicolumn{1}{l|}{{0.005/0.003 (0.945)}} &
  \multicolumn{1}{l|}{{0.104/0.077 (1.021)}} &
  {0.037/0.027 (0.920)} \\
\multicolumn{1}{l|}{{UniEval}} &
  \multicolumn{1}{l|}{{\underline{0.648}/0.512 (1.830)}} &
  \multicolumn{1}{l|}{{\textbf{0.571}/0.447 (1.079)}} &
  \multicolumn{1}{l|}{{0.539/0.404 (1.675)}} &
  \multicolumn{1}{l|}{{0.504/0.481 (1.755)}} &
  \multicolumn{1}{l|}{{\underline{0.627}/0.483 (1.443)}} &
  \multicolumn{1}{l|}{{0.530/\underline{0.475} (1.635)}} &
  \multicolumn{1}{l|}{{\underline{0.679}/\underline{0.535} (1.555)}} &
  {\underline{0.585}/0.477 (1.567)} \\ \hline
{\textbf{LLM based (GPT 3.5-Turbo)}} &
  {} &
  {} &
  {} &
  {} &
  {} &
  {} &
  {} &
  {} \\ \hline
\multicolumn{1}{l|}{{G-Eval (GPT 3.5-Turbo)}} &
  \multicolumn{1}{l|}{{0.441/0.342 (1.058)}} &
  \multicolumn{1}{l|}{{0.451/0.372 (1.032)}} &
  \multicolumn{1}{l|}{{0.378/0.289 (\underline{0.872})}} &
  \multicolumn{1}{l|}{{0.388/0.354 (0.888)}} &
  \multicolumn{1}{l|}{{0.379/0.288 (\underline{0.809})}} &
  \multicolumn{1}{l|}{{0.402/0.304 (\underline{0.679})}} &
  \multicolumn{1}{l|}{{0.346/0.262 (1.024)}} &
  {0.398/0.316 (\underline{0.909})} \\
\multicolumn{1}{l|}{{CheckEval (GPT 3.5-Turbo)}} &
  \multicolumn{1}{l|}{{0.366/0.329 (1.942)}} &
  \multicolumn{1}{l|}{{0.335/0.242 (1.992)}} &
  \multicolumn{1}{l|}{{0.283/0.258 (1.952)}} &
  \multicolumn{1}{l|}{{-0.061/-0.055 (1.624)}} &
  \multicolumn{1}{l|}{{0.401/0.304 (1.849)}} &
  \multicolumn{1}{l|}{{0.487/0.418 (1.534)}} &
  \multicolumn{1}{l|}{{0.331/0.288 (1.729)}} &
  {0.306/0.255 (1.803)} \\
\multicolumn{1}{l|}{{InteractEval}} &
  \multicolumn{1}{l|}{{0.431/0.365 (1.324)}} &
  \multicolumn{1}{l|}{{0.392/0.349 (1.921)}} &
  \multicolumn{1}{l|}{{0.486/0.416 (1.571)}} &
  \multicolumn{1}{l|}{{0.470/0.413 (1.816)}} &
  \multicolumn{1}{l|}{{0.480/0.407 (1.614)}} &
  \multicolumn{1}{l|}{{0.479/0.423 (1.643)}} &
  \multicolumn{1}{l|}{{0.392/0.346 (1.568)}} &
  {0.447/0.388 (1.637)} \\ \hline
{\textbf{LLM based (GPT-4)}} &
  {} &
  {} &
  {} &
  {} &
  {} &
  {} &
  {} &
  {} \\ \hline
\multicolumn{1}{l|}{{G-Eval (GPT 4)}} &
  \multicolumn{1}{l|}{{0.622/0.501 (0.918)}} &
  \multicolumn{1}{l|}{{0.562/\underline{0.458} (0.944)}} &
  \multicolumn{1}{l|}{{\underline{0.542}/\underline{0.455} (1.004)}} &
  \multicolumn{1}{l|}{{\underline{0.626}/\underline{0.491} (0.978)}} &
  \multicolumn{1}{l|}{{0.555/0.431 (1.174)}} &
  \multicolumn{1}{l|}{{\underline{0.560}/0.443 (1.041)}} &
  \multicolumn{1}{l|}{{0.610/0.479 (1.036)}} &
  {0.582/0.465 (1.014)} \\
\multicolumn{1}{l|}{{CheckEval (GPT 4)}} &
  \multicolumn{1}{l|}{{0.645/\underline{0.538} (1.044)}} &
  \multicolumn{1}{l|}{{0.557/0.457 (1.224)}} &
  \multicolumn{1}{l|}{{0.510/0.421 (0.980)}} &
  \multicolumn{1}{l|}{{0.502/0.453 (\textbf{0.638})}} &
  \multicolumn{1}{l|}{{0.610/\underline{0.527} (1.373)}} &
  \multicolumn{1}{l|}{{0.542/0.441 (1.120)}} &
  \multicolumn{1}{l|}{{0.635/0.527 (\textbf{0.806})}} &
  {0.572/\underline{0.481} (1.026)} \\
\multicolumn{1}{l|}{{InteractEval}} &
  \multicolumn{1}{l|}{{\textbf{0.691}/\textbf{0.605} (\textbf{0.547})}} &
  \multicolumn{1}{l|}{{\underline{0.567}/\textbf{0.460} (\textbf{0.928})}} &
  \multicolumn{1}{l|}{{\textbf{0.554}/\textbf{0.470} (0.982)}} &
  \multicolumn{1}{l|}{{\textbf{0.631}/\textbf{0.505} (\underline{0.855})}} &
  \multicolumn{1}{l|}{{\textbf{0.644}/\textbf{0.543} (\textbf{0.641})}} &
  \multicolumn{1}{l|}{{\textbf{0.581}/\textbf{0.499} (\textbf{0.662})}} &
  \multicolumn{1}{l|}{{\textbf{0.685}/\textbf{0.581} (\underline{0.886})}} &
  {\textbf{0.622}/\textbf{0.523} (\textbf{0.786})} \\ \hline
\end{tabular}%
}
\end{table*}

\end{document}